%% file: bare_jrnl_compsoc.tex
\documentclass[10pt,journal,compsoc]{IEEEtran}
\ifCLASSOPTIONcompsoc
\else
  \usepackage{cite}
\fi
\ifCLASSINFOpdf
\else
\fi
\usepackage{ragged2e}
\usepackage{xspace}
\usepackage{subfigure}
\usepackage{graphicx}
\usepackage{multirow}
\usepackage{hhline}
\usepackage{comment}
\usepackage{bigstrut}
\usepackage[dvipsnames]{xcolor}
\usepackage[sort&compress,square,comma,numbers]{natbib}
\usepackage[colorlinks=false, linkcolor=blue, citecolor=blue, bookmarks=true,pagebackref=false]{hyperref}
\usepackage{hyphenat}
\usepackage{longtable}
\usepackage{subcaption}
\usepackage{float}
\usepackage{fancybox}
\usepackage[flushleft]{threeparttable}
\usepackage{array,booktabs}
\usepackage{amsmath}
\usepackage{outlines}
\usepackage[linesnumbered,ruled]{algorithm2e}
\usepackage[misc]{ifsym}
\usepackage{arydshln}
\usepackage{balance}
\usepackage{amsmath}
\usepackage{MnSymbol}
\usepackage{wasysym}
\usepackage{enumitem}
\usepackage{times}
\usepackage{fancyhdr}
\usepackage{courier}
\usepackage{booktabs}
\usepackage{pifont}
\usepackage{color,soul} %Allows text highlighting
\usepackage{tcolorbox}
\usepackage{ragged2e}
\usepackage[justification=centering]{caption}
\usepackage{natbib}
\usepackage{tikz}
\usepackage{capt-of}
\usepackage{graphicx}
\usepackage{hyperref}
\usepackage{float}
\usepackage{tablefootnote}
\usepackage{natbib}

\newtcolorbox{mybox}[2][]{
top=0.15in,left=4pt,right=4pt,bottom=4pt,
fonttitle=\bfseries,
colbacktitle=gray,
colback=gray!5,
colframe=gray!40!black,
enhanced,
attach boxed title to top left={xshift=1.5em,yshift=-\tcboxedtitleheight/2},
boxed title style={size=small},
drop shadow={black!50!white},
title=#2,#1}

\usepackage{soul}
\definecolor{hzcolor}{RGB}{10, 186, 181}
\setstcolor{red}

\begin{document}
%\title{EfficientDL: An Automated Performance Estimation
%and Recommendation Framework for Deep Learning}

\title{Efficient Deep Learning Board: Training Feedback Is Not All You Need}

%~\IEEEmembership{Member,~IEEE,}
\author{Lina~Gong,
        Qi~Gao, Peng~Li, Mingqiang~Wei, \textit{Senior Member}, \textit{IEEE}, and Fei~ Wu, \textit{Senior Member}, \textit{IEEE}% <-this % stops a space

        \thanks{Lina Gong, Qi Gao, Peng Li and Mingqiang Wei are with the School of Computer Science and Technology, Nanjing University of Aeronautics and Astronautics, Nanjing, China (e-mail: {gonglina,gaoqiCuCu, pengl, mqwei}@nuaa.edu.cn. }
          \thanks{Fei Wu is with the School of Computer Science and Technology, Zhejiang University, Hangzhou, China (e-mail: wufei@zju.edu.cn ).}

%\thanks{Manuscript received October 19, 2024; revised October 26, 2024.}
}


\markboth{Journal of \LaTeX\ Class Files,~Vol.~14, No.~8, August~2015}%
{Shell \MakeLowercase{\textit{et al.}}: Bare Demo of IEEEtran.cls for Computer Society Journals}



\IEEEtitleabstractindextext{%
\begin{abstract}
\input{01-Abstract.tex}
\end{abstract}

}

\maketitle

\IEEEdisplaynontitleabstractindextext

% \fi
\IEEEpeerreviewmaketitle

\IEEEraisesectionheading{\section{Introduction}\label{sec:introduction}}

\input{02-introduction.tex}

\input{03-relatedwork.tex}
\input{05-DeepImageBenchdata}

\input{04-EfficientDL.tex}
\input{06-Experiment.tex}
\input{07-Conclusion.tex}
\ifCLASSOPTIONcaptionsoff
  \newpage
\fi



%\bibliographystyle{IEEEtranSN}
\bibliographystyle{IEEEtran}
%\begin{footnotesize}
%\balance
\bibliography{reference}
%\end{footnotesize}
%\begin{thebibliography}{1}


%\end{thebibliography}


\begin{IEEEbiography}[{\includegraphics[width=1in,height=1.25in,clip,keepaspectratio]{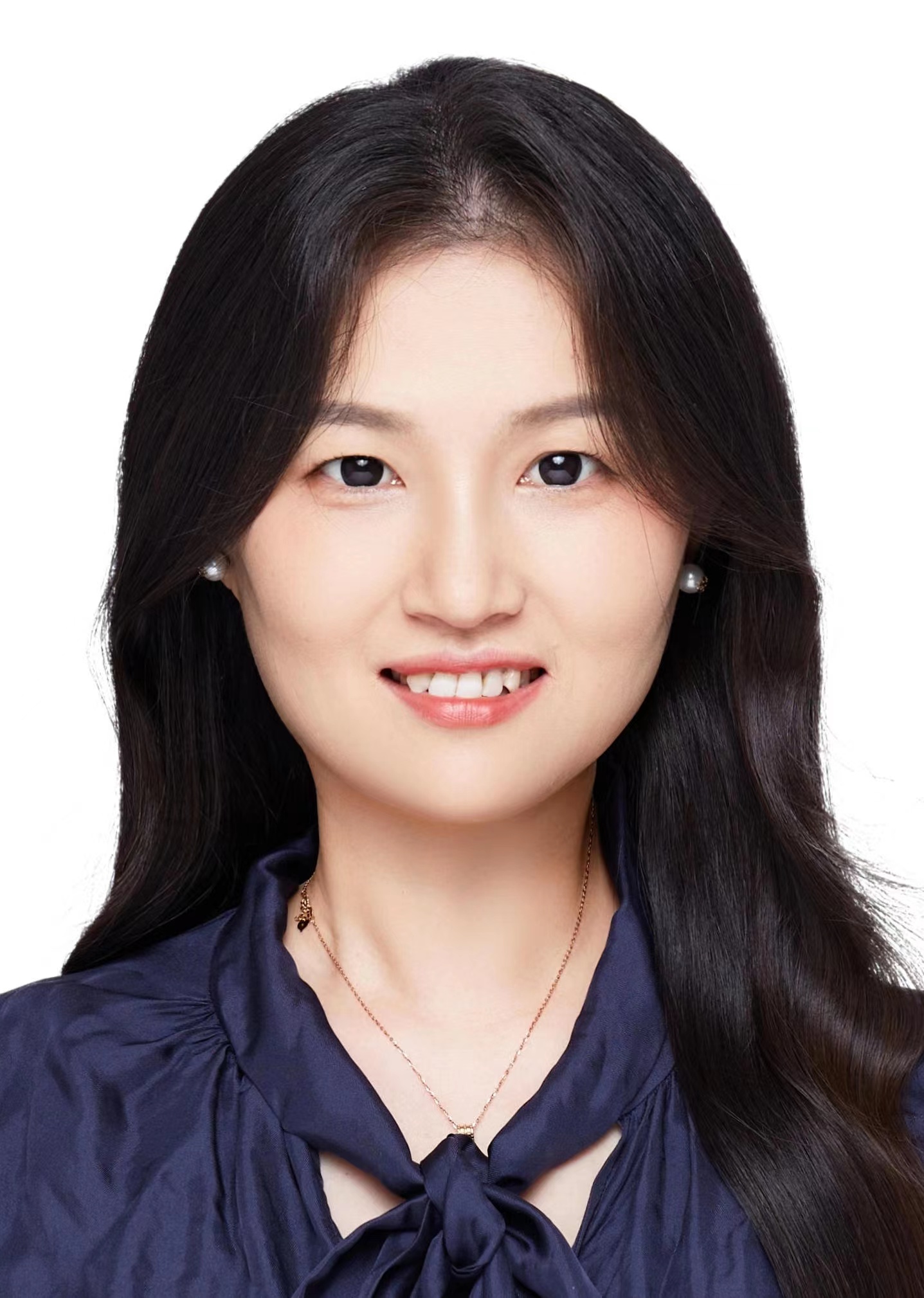}}]{Lina Gong}
is currently an associate professor in the College    of    Computer    Science    and    Technology,   Nanjing   University   of   Aeronautics   and   Astronautics, China.  She received her Ph.D. degree in the Computer software and theory from China University of Mining and Technology, China, her BE degree in Software Engineering from China University of Petroleum, China. She also studied as a visitor one year in the Software Analysis and Intelligence Lab (SAIL), School of Computing, Queen’s University, Canada. Her research interests include machine learning, software analysis, software testing and mining software repositories. More information at: \url{https://lina-gong.github.io/}.
\end{IEEEbiography}.

\begin{IEEEbiography}[{\includegraphics[width=1in,height=1.25in,clip,keepaspectratio]{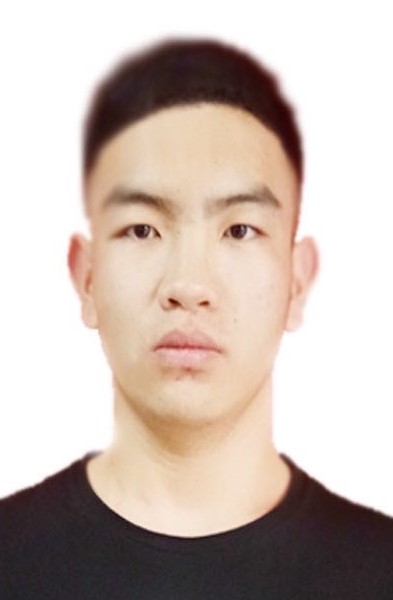}}]
{Qi Gao} is currently a Master’s student in Software Engineering at the School of Computer Science and Technology, Nanjing University of Aeronautics and Astronautics. He received his Bachelor’s degree in Computer Science and Technology from Shanxi University. His research interests include image classification and neural architecture search.
\end{IEEEbiography}

\begin{IEEEbiography}[{\includegraphics[width=1in,height=1.25in,clip,keepaspectratio]{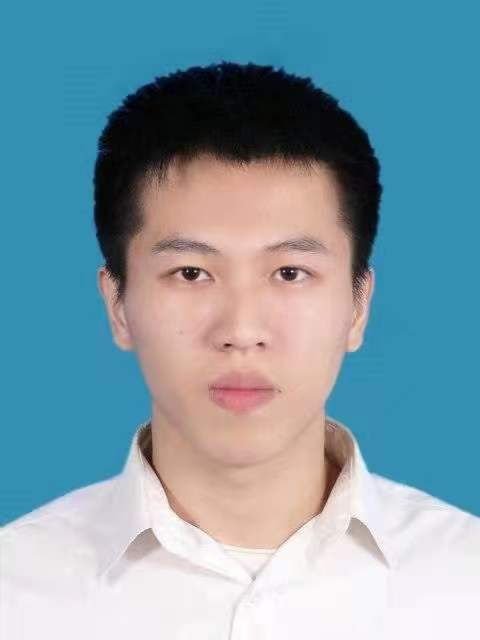}}]{Peng Li}
is a Ph.D. candidate at the School of Computer Science and Technology, Nanjing University of Aeronautics and Astronautics, China. He has published several papers on IEEE TIP, AAAI, PG, etc. His research interests include deep learning, image processing, and computer vision.
\end{IEEEbiography}

%\begin{IEEEbiography}[{\includegraphics[width=1in,height=1.25in,clip,keepaspectratio]{figures/Yongwei_Wang.png}}]{Yongwei Wang}
%is a currently a Researcher at Shanghai Institute for Advanced Study and College of Computer Science, Zhejiang University in China. Previously, he received his Ph.D. from the University of British Columbia in Canada in 2021, and was a Research Fellow at Nanyang Technological University in Singapore from 2021 to 2023. His research interests include generative AI, AI security and multimedia forensics.  
%\end{IEEEbiography}

\begin{IEEEbiography}[{\includegraphics[width=1in,height=1.25in,clip,keepaspectratio]{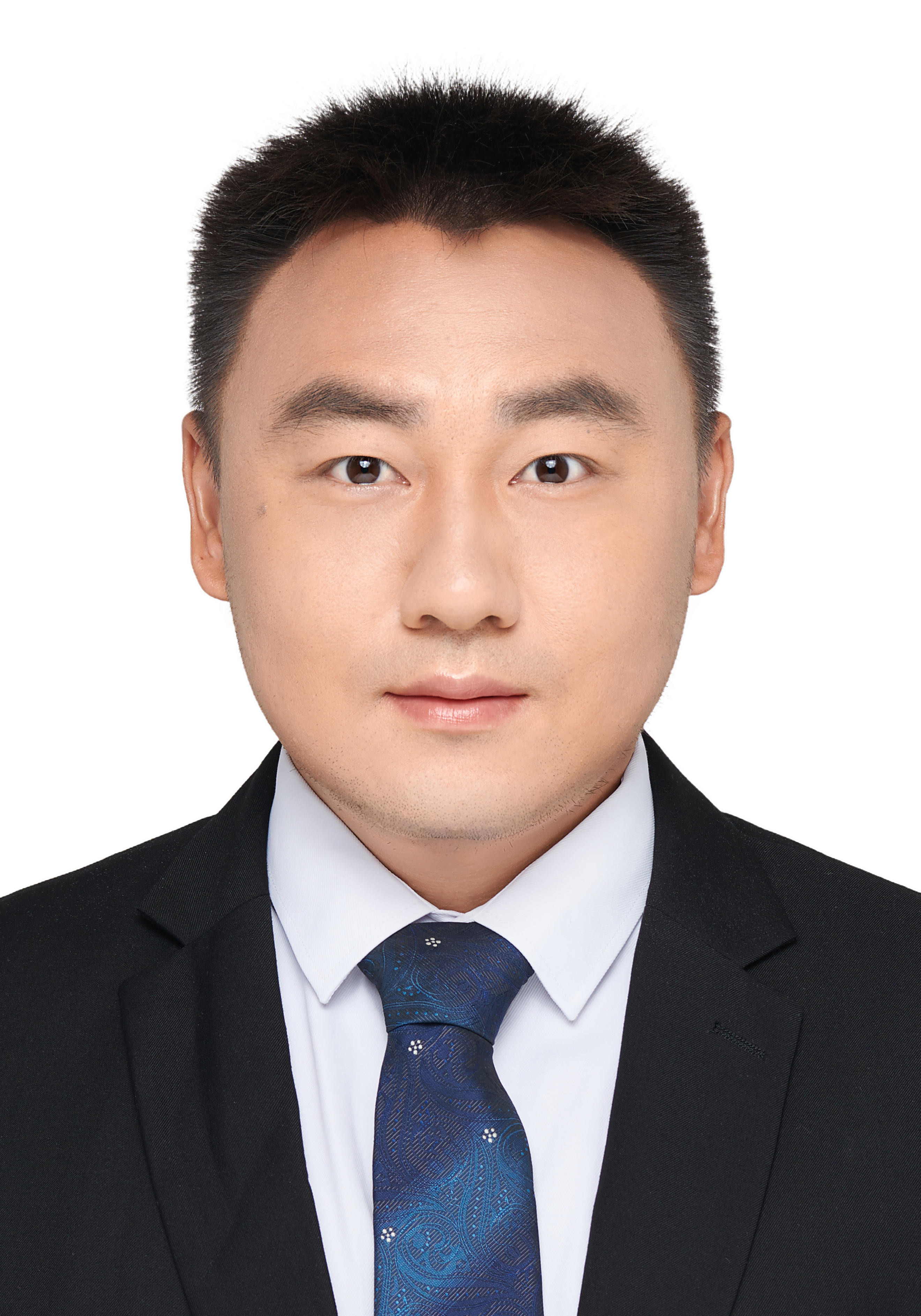}}]{Mingqiang Wei}
	received his Ph.D degree (2014) in Computer Science and Engineering from the Chinese University of Hong Kong (CUHK). He is a full Professor at the School of Computer Science and Technology, Nanjing University of Aeronautics and Astronautics (NUAA).  He was a recipient of the CUHK Young Scholar Thesis Awards in 2014. He is now an Associate Editor for ACM TOMM, The Visual Computer (TVC), and a leading Guest Editor for IEEE Transactions on Multimedia. He has published
150 research publications, including TPAMI, SIGGRAPH, TVCG, CVPR, ICCV, et al. His research interests focus on 3D vision, computer graphics, and deep learning.
\end{IEEEbiography}

\begin{IEEEbiography}[{\includegraphics[width=1in,height=1.25in,clip,keepaspectratio]{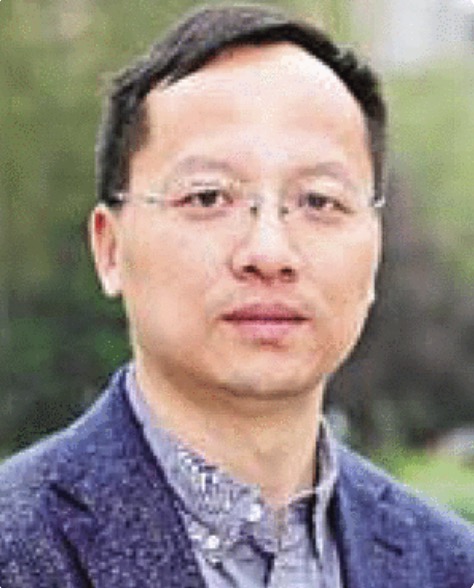}}]{Fei Wu}
(Senior Member, IEEE) received the PhD degree from Zhejiang University, Hangzhou, China. He was a visiting scholar with the prof. B. Yu's Group, University of California at Berkeley, Berkeley, from 2009 to 2010. He is currently a full professor with the College of Computer Science and Technology, Zhejiang University. His current research interests include multimedia retrieval, sparse representation, and machine learning. 
\end{IEEEbiography}


%\newpage


%\vfill

\newpage
\onecolumn
\appendices
\section{ImageClassEval Dataset with Designed System components}

ImageClassEval is specifically designed to support automated performance evaluation and component
recommendation. Table~\ref{tab:model_architecture} presents the designed system components and the range of each system component. Meanwhile, we compare our ImageClassEval with other datasets, including NASBench-101, NASBench-201, LCBench and HPOBench.

% \small
\begin{longtable}
{p{0.08\textwidth}p{0.08\textwidth}p{0.045\textwidth}p{0.14\textwidth}p{0.045\textwidth}p{0.05\textwidth}p{0.045\textwidth}p{0.05\textwidth}p{0.045\textwidth}p{0.05\textwidth}p{0.045\textwidth}p{0.05\textwidth}}
\caption{System Components included in different benchmarks}
\label{tab:model_architecture}\\
\toprule
\textbf{Dimension} & \textbf{Component} & 
\multicolumn{2}{c}{\textbf{ImageClassEval}}  & \multicolumn{2}{c}{\textbf{NAS-Bench-101}}& 
\multicolumn{2}{c}{\textbf{NAS-Bench-201}} & 
\multicolumn{2}{c}{\textbf{LCBench} } & 
\multicolumn{2}{c}{\textbf{HPOBench}} \\
 &  & Yes or No & range & Yes or No & range & Yes or No & range & Yes or No & range & Yes or No & range\\
\midrule
\endfirsthead
\toprule
\textbf{Dimension} & \textbf{Component}&
\multicolumn{2}{c}{\textbf{ImageClassEval}} & \multicolumn{2}{c}{\textbf{NAS-Bench-101}}& 
\multicolumn{2}{c}{\textbf{NAS-Bench-201}} & 
\multicolumn{2}{c}{\textbf{LCBench} } & 
\multicolumn{2}{c}{\textbf{HPOBench}} \\

 &  & Yes or No & range & Yes or No & range & Yes or No & range & Yes or No & range & Yes or No & range\\
\midrule
\endhead
Model architecture & Normali-zation layer & $\surd$ & \{Batch Normalization, Spectral Normalization, Group Normalization, Layer Normalization, Conditional Batch Normalization, Attentive Normalization, LayerScale, Weight Standardization, Local Response Normalization\} & $\times$ & - & $\times$ & - & $\times$ & - & $\times$ & -
\\
\cline{2-12}

 & Initializ-ation & $\surd$ & \{Kaiming Initialization, Xavier Initialization, Fixup
Initialization, LSUV Initialization\} & $\times$ & - & $\times$ & - & $\times$ & - & $\times$ & -
\\
\cline{2-12}
 & Convolu-tion & $\surd$ & \{Depthwise Convolution, Grouped Convolution, Pointwise Convolution, 3x3 Convolution, Selective Kernel Convolution, 1x1 Convolution, Depthwise Separable Convolution, MixConv, Spatially Separable Convolution, Gated Convolution\} & $\surd$ & \{3×3 convolution, 1×1 convolution\} & $\surd$ & \{3×3 convolution, 1×1 convolution\} & $\times$ & - & $\times$ & -
 \\
\cline{2-12}
 & Skip Connection & $\surd$ & \{Residual Connection, Concatenated Skip Connection, Zero-padded Shortcut Connection, Deactivable Skip Connection\} & $\times$ & - & $\surd$ & \{None, skip connection\} & $\times$ & - & $\times$ & -
 \\
\cline{2-12}
 & Activation Function & $\surd$ & \{GLU, ReLU, CReLU, Leak ReLU, Tanh Activation, GELU, PReLU, Sigmoid, Hard Swish, Swish, Sigmoid Activation, Sigmoid Linear Unit, Softplus\} & $\times$ & - & $\times$ & - & $\times$ & - & $\times$ & -
 \\
\cline{2-12}
 & Pooling Operation & $\surd$ & \{Spatial Pyramid Pooling, Average Pooling, Generalized Mean Pooling, Global Average Pooling, Max Pooling, Center Pooling\} & $\surd$ & \{None, 3×3 max-pooling\} & $\surd$ & \{None, 3×3 avg pooling\} & $\times$ & - & $\times$ & -
 \\
\cline{2-12}
 
 & Feedforward Network & $\surd$ & \{Dense Connections, Linear Layer, Position-Wise Feed-Forward Layer, Feedforward Network, Affine Operator\} & $\times$ & - & $\times$ & - & $\times$ & - & $\times$ & -
 \\
\cline{2-12}
 & Attention Mechanism & $\surd$ & \{Scaled Dot-Product Attention, Recurrent models of visual attention, Fast Attention Via Positive Orthogonal Random Features, Recurrent models of visual attention, linear attention mechanism, Pooling Attention, Class Attention, Channel Attention, Dilated Neighborhood Attention, Multi-axis Attention, Channel-wise Soft Attention, Dilated Sliding Window Attention, Global and Sliding Window Attention, Sliding Window Attention, Multi-Head Attention, Restricted Self-Attention\} & $\times$ & - & $\times$ & - & $\times$ & - & $\times$ & -
 \\
\cline{2-12}

 & Output Function & $\surd$ & \{Softmax, Heatmap, Mixture of Logistic Distributions, Adaptive Softmax, Extreme Value Machine, Sparsemax, PAFs\} & $\times$ & - & $\times$ & - & $\times$ & - & $\times$ & -
 \\
\cline{1-12}
Training Optimization & Learning Rate Schedule & $\surd$ & \{Cosine Annealing, Linear Warmup With Cosine Annealing, Linear Warmup With Linear Decay, Exponential Decay, Cosine Power Annealing, Log Decay, Linear Warmup, Polynomial Rate Decay\} & $\times$ & - & $\times$ & - & $\times$ & - & $\times$ & -
 \\
 \cline{2-12}
 & Optimization algorithm & $\surd$ & \{AdamW, SGD, RMSProp, LAMB, AdamP, AdaGrad, Adam, LARS optimizer, Nesterov momentum optimizer, SGD with Momentum\} & $\times$ & - & $\times$ & - & $\surd$ & \{SGD, Adam\} & $\times$ & -
 \\
 \cline{2-12}
 & Size of parameter & $\surd$ & [0.18M, 632M] & $\times$ & - & $\times$ & - & $\times$ & - & $\times$ & -\\
 \cline{2-12}
 & Batch size & $\surd$ & [32, 8192] & $\times$ & - & $\times$ & - & $\surd$ & [16, 512] & $\surd$ & [4, 256]\\
 \cline{2-12}
 & Learning rate & $\surd$ & [0.0000025, 4.8] & $\times$ & - & $\times$ & - & $\surd$ & [0.0001, 0.1] & $\surd$ & [0.00001, 1]
 \\
 \cline{2-12}
 & Epochs & $\surd$ & [20, 5000] & $\times$ & - & $\times$ & - & $\times$ & - & $\surd$ & [3, 243]
 \\
 \hline
Regulariz-ation and Generalization & Regulariz-ation & $\surd$ & \{Dropout, Label Smoothing, Weight Decay, R1 Regularization, L1 Regularization, L2 Regularization, DropBlock\} & $\times$ & - & $\times$ & - & $\times$ & - & $\times$ & -
 \\
 \cline{2-12}
& Data Augmentation & $\surd$ & \{random horizontal flip, random vertical flip, random flip, random translation, random rotation, random resized crop, center crop, random crop, colorjitter, random Lighting Noise, saturation delta, random brightness, solarization, autoaugment,randaugment,convert to gray scale, random scale, gaussian blur, mixup, cutout, random erasing, cutmix, inception crop\} & $\times$ & - & $\times$ & - & $\surd$ & \{standard, mixup\} & $\times$ & -
 \\
\cline{1-12}
Framework & Framework & $\surd$ & \{Caffe, Caffe2, tensorflow, PyTorch\} & $\times$ & - & $\times$ & - & $\times$ & - & $\times$ & -
 \\
 \hline
Data & Size of training set & $\surd$ & [2360, 1803460] & $\times$ & - & $\times$ & - & $\times$ & - & $\times$ & -
 \\
\cline{2-12}
 & Size of testing set & $\surd$ & [238, 328500] & $\times$ & - & $\times$ & - & $\times$ & - & $\times$ & -
 \\
\cline{2-12}
 & Input Length & $\surd$ & [16, 512] & $\times$ & - & $\times$ & - & $\times$ & - & $\times$ & -
 \\
\cline{2-12}
 & Output Length & $\surd$ & [5, 5089] & $\times$ & - & $\times$ & - & $\times$ & - & $\times$ & -
 \\
\cline{2-12}
 & Cosine similarity & $\surd$ & [0.0000089407, 0.09754324] & $\times$ & - & $\times$ & - & $\times$ & - & $\times$ & -
 \\
\cline{2-12}
 & Jensen-Shannon & $\surd$ & [0.05358259, 0.408972877] & $\times$ & - & $\times$ & - & $\times$ & - & $\times$ & -
 \\
\cline{2-12}
 & L2 distance & $\surd$ & [0.147056907, 0.151814908] & $\times$ & - & $\times$ & - & $\times$ & - & $\times$ & -\\
\cline{1-12}
 
Hardware & Type of GPU & $\surd$ & \{TPU-v3, TPU-v2, TPUv4, NVIDIA TESLA K80, NVIDIA V100, NVIDIA A100, NVIDIA A800, NVIDIA GeForce GTX 1080 Ti, NVIDIA GeForce RTX 2080 Ti, NVIDIA GeForce RTX 3090, NVIDIA GTX 580, NVIDIA GTX980, Nvidia RTX 3070, Tesla P100, Quadro RTX 8000, RTX A5000, Nvidia Tesla K40, NVIDIA M40, Titan Xp GPUs, Titan X GPU, Nvidia P40 GPUs, Tesla T4 GPU, cpu\} & $\times$ & - & $\times$ & - & $\surd$ & \{Intel Xeon Gold 6242 CPUs\} & $\surd$ & \{GeForce GTX 1080 Ti\} 
\\
\cline{2-12}
 & Number of GPU & $\surd$ & [1, 60] & $\times$ & - & $\times$ & - & $\times$ & - & $\times$ & -\\
 \hline
\end{longtable}
% \end{singlecolumn}
\onecolumn
\section{\texorpdfstring{$\alpha$-$\beta$ BO Search Algorithm}{Alpha-Beta BO Search Algorithm}}
{\ensuremath{\alpha \beta}-BO search Algorithm is specifically designed for optimization problems within large search spaces, as shown in Algorithm \ref{alg:bo_algorithm}. Lines 1-2 define the algorithm’s input and output parameters, including the surrogate model for Bayesian optimization, referred to as the GP model, the initial dataset $\mathcal{D}_0$, the maximum number of iterations $t$, and the search space $\mathcal{X}$. In Line 3, the GP model is initialized using the dataset $\mathcal{D}_0$. Lines 4-14 describe the iterative process of the algorithm, which involves selecting the next sampling point $x{i+1}$, computing the corresponding observation $y_{i+1}$, updating the GP model, and adjusting the random search probability $\Omega$.

\begin{algorithm}[H]\label{alg:bo_algorithm}
\SetAlgoLined 
\DontPrintSemicolon % 不打印分号
\KwIn{%
    GP model, initial data $\mathcal{D}_0 = \{ (x_i, y_i) \}_{i=1,...N}$, upper limits $t$ , search space $\mathcal{X}$
}
\KwOut{Optimal solution $x^*$}
\BlankLine
Initialize the GP model with data $\mathcal{D}$\;

\For{$i = 0, 1, \ldots, t$}{
    Generate a random number $\xi$ in the range $[0, 1]$ \\
    \If{$\xi \leq \Omega$}{
        Perform random sampling to generate the next sample point $x_{\text{i+1}}$\;
    }
    \If {$\xi > \Omega$}{
        $x_{i+1}=\underset{x\in\mathcal{X}}{\operatorname*{argmax}} \; \alpha_{t}^{\gamma \text{EI}}(x) $
    }
    Query the objective function to obtain $y_{\text{i+1}}$ \\
    Augment $\mathcal{D}_{i+1} = \{\mathcal{D}_{i}, (x_{i+1},y_{i+1})\}$ and update statistical model GP\\
    Update the $\Omega$ according to equation\eqref{eq:Om}
}
\caption{$\alpha\beta$-BO Algorithm}
\label{algo:bubbleSort}
\end{algorithm}
}
\end{document}

%% file: 01-Abstract.tex
\justifying
Current automatic deep learning (i.e., AutoDL) frameworks rely on training feedback from actual runs, which often hinder their ability to provide quick and clear performance predictions for selecting suitable DL systems. To address this issue, we propose EfficientDL, an innovative deep learning board designed for automatic performance prediction and component recommendation. EfficientDL can quickly and precisely recommend twenty-seven system components and predict the performance of DL models without requiring any training feedback. 
% The magic of no training feedback comes from our proposed static performance prediction solution, which removes the dependency on actually running parameterized models during the optimization search. with our improved $\alpha\beta$-BO search 
The magic of no training feedback comes from our proposed comprehensive, multi-dimensional, fine-grained system component dataset, which enables us to develop a static performance prediction model and comprehensive optimized component recommendation algorithm (i.e., $\alpha\beta$-BO search), removing the dependency on actually running parameterized models during the traditional optimization search process.
The simplicity and power of EfficientDL stem from its compatibility with most DL models. 
% For example, EfficientDL operates seamlessly with mainstream models such as ResNet50, MobileNetV3, EfficientNet-B0, MaxViT-T, Swin-B, and DaViT-T, demonstrating competitive performance compared to existing AutoML tools for image classification. 
For example, EfficientDL operates seamlessly with mainstream models such as ResNet50, MobileNetV3, EfficientNet-B0, MaxViT-T, Swin-B, and DaViT-T, bringing competitive performance improvements. 
Besides, experimental results on the CIFAR-10 dataset reveal that EfficientDL outperforms existing AutoML tools in both accuracy and efficiency (approximately 20 times faster along with 1.31\% Top-1 accuracy improvement than the cutting-edge methods).
Source code, pretrained models, and datasets are available at \url{https://github.com/OpenSELab/EfficientDL}.

%% file: 02-introduction.tex
\noindent \IEEEPARstart{D}{eep} learning is showcasing its immense potential across various domains, including computer vision and natural language processing. However, \textbf{can practitioners truly harness the full power of deep learning models?} Researchers and practitioners often face the challenge of investing substantial time and computational resources to manually select suitable model architectures, tune hyperparameters (such as learning rate, batch size, and number of epochs), and augment data to align with the specific characteristics of their datasets. As a result, achieving optimal performance with deep learning models can be particularly daunting for beginners.

%However, in practice, achieving optimal performance from deep learning models in specific domains is not a straightforward endeavor. 

%Researchers and practitioners  must invest significant time and computational resources to manually select appropriate model architectures, tune various hyperparameters (e.g., learning rate, batch size, and epoch), and augment data to adapt to the characteristics of their specific datasets to achieve the desired performance.

 \begin{figure}[htbp]
    \centering
    \includegraphics[width=1.0\linewidth, scale=1.0]{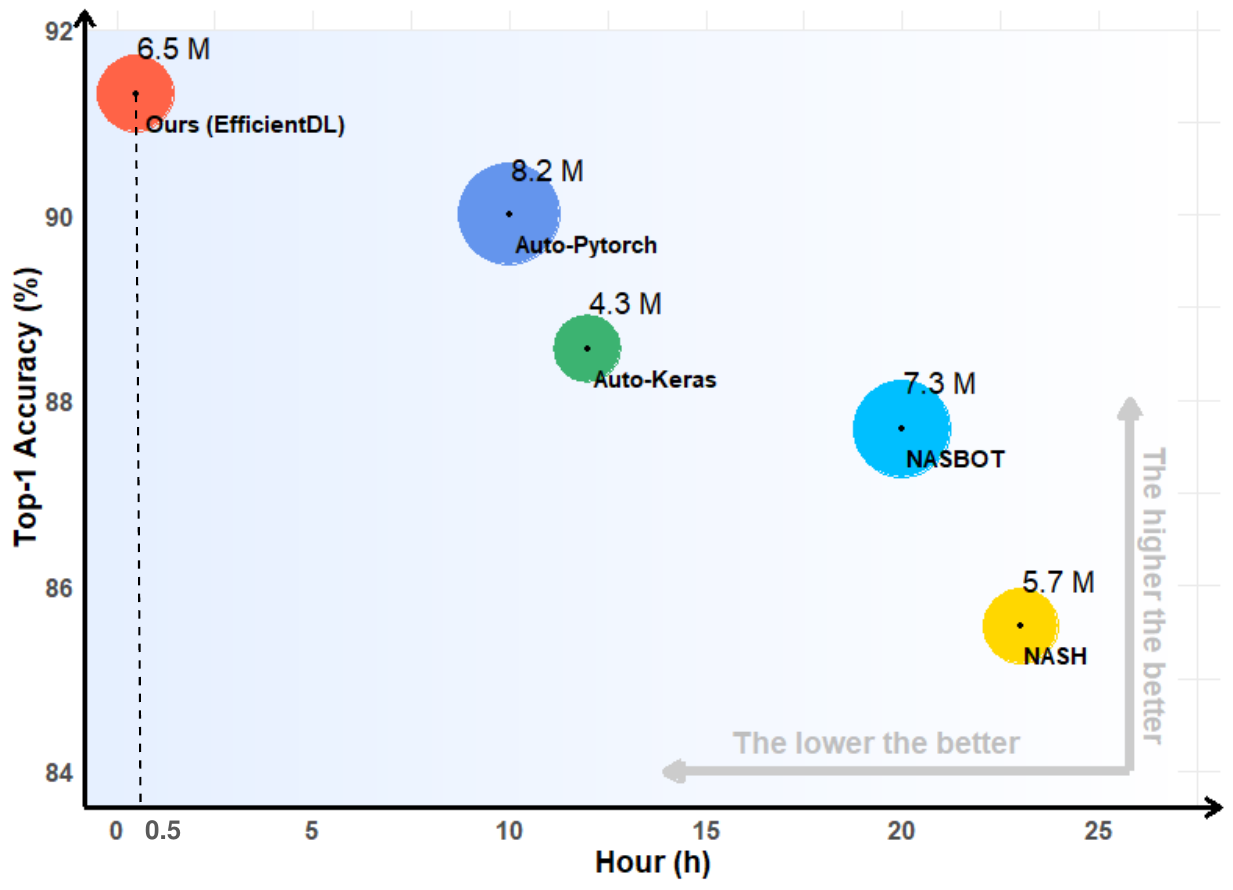}
    \captionsetup{justification=justified}
    \caption{The comparison between EfficientDL and other AutoDL frameworks on the CIFAR-10 dataset. Compared to other AutoDL frameworks that take at least 10 hours to discover an optimal model with 8.2M parameters and achieve 90\% Top-1 accuracy on CIFAR-10, our EfficientDL identifies an optimal model with 6.5M parameters in just 0.5 hours, achieving a higher performance of 91.31\%.}
    \label{fig:performence}
\end{figure}

Recently, automatic machine learning (i.e., AutoML) techniques have emerged to streamline the design of deep learning systems, including tools like AutoML-CFBO~\cite{nicolonips2018}, AutoKeras~\cite{haifeng2023jml}, and AutoFormer~\cite{chen2021iccv}. These techniques enhance the performance of machine learning models by automatically discovering optimal model architectures and hyperparameters for specific tasks. As a result, AutoML has proven to be invaluable for deep learning developers, enabling them to more easily apply deep learning to their particular applications.

%For example, Nicolo et al.~\cite{nicolonips2018} propose an AutoML technique combining ideas from collaborative filtering and Bayesian optimization to tune the data prepossessing strategies and machine learning models. Jin et al.~\cite{haifeng2023jml} develop AutoKeras to automatically select model architectures and hyperparameters based on Keras and TensorFlow. Chen et al.~\cite{chen2021iccv} propose AutoFormer for vision transformer search through entangling the weights of different blocks in the same layers during supernet training.  

%These AutoML techniques have been fruitful in enabling DL developers to more conveniently apply deep learning to their specific tasks. 

However, the optimization rules of these AutoML methods rely on the models' training feedback under given parameters. For example, upon receiving a new set of parameters, LEAF~\cite{Jason2019GECCO} requires training the deep learning model for three epochs under the current parameters to assess their impact, while Auto-Pytorch~\cite{Lucas2021TPMI} utilizes the deep learning model trained for fifty epochs for evaluation. 
Consequently, the time and resource savings from AutoML methods are also limited, especially when dealing with large training datasets.  For example, as illustrated in Fig.~\ref{fig:performence}, Auto-Pytorch took 10 hours to discover optimal model architectures and hyperparameters for image classification task on CIFAR-10. 

%However, the optimization rules of these AutoML methods depend heavily on the training feedback from models under specific parameters. For instance, LEAF~\cite{Jason2019GECCO} requires training the deep learning model for three epochs with a new set of parameters to evaluate their impact, while Auto-PyTorch~\cite{Lucas2021TPMI} assesses performance based on models trained for fifty epochs. Consequently, the time and resource savings offered by AutoML methods are limited, particularly when working with large training datasets.

%Moreover, current AutoML methods mainly focus on a single constrained search space, typically involving fixed model architectures and specific hyperparameters within a limited range. For instance, ProxylessNAS~\cite{cai2019ICLR} is designed for the automated construction and optimization of convolutional neural networks, concentrating on key parameters such as kernel size, number of channels, network depth, and skip connections. Similarly, LiteTransformerSearch~\cite{Mojan2022NeurIPS} automates the design and optimization of architectures within the GPT family, emphasizing parameters like layer count, model dimension, adaptive embedding size, feedforward network dimensions within transformer layers, and the number of attention heads per layer. Due to the restricted nature of these search spaces, such methods may struggle to reach their full potential when tackling more complex tasks, particularly those that require a holistic consideration of multiple factors, data variations, and diverse hardware environments.

Inspired by current shortcomings that rely on training feedback from actual runs in AutoML, we explore into the following two questions:

\begin{enumerate}
    \item Can AutoML be endowed with the ability to foresee model performance, thereby circumventing the need for performance evaluation?
    \item How can we ensure the accuracy of performance forecasts?
\end{enumerate}

We argue that the resolution of the above questions pivots on \textbf{the construction of a comprehensive, multi-dimensional, fine-grained system component dataset.}
Such a dataset serves a dual purpose:
1) it facilitates the development of a static performance prediction model in a data-driven manner, thereby endowing the AutoML methods with predictive capabilities. 2) The diversity and richness of the dataset contribute to the robustness and reliability of the performance prediction model.
However, current AutoML methods mainly focus on a single constrained search space, typically involving fixed model architectures and specific hyperparameters within a limited range. For instance, ProxylessNAS~\cite{cai2019ICLR} is designed for the automated construction and optimization of convolutional neural networks, concentrating on key parameters such as kernel size, number of channels, network depth, and skip connections. Similarly, LiteTransformerSearch~\cite{Mojan2022NeurIPS} automates the design and optimization of architectures within the GPT family, emphasizing parameters like layer count, model dimension, adaptive embedding size, feedforward network dimensions within transformer layers, and the number of attention heads per layer. Such benchmarks may struggle to reach their full potential when tackling more complex tasks, particularly those that require a holistic consideration of multiple factors, data variations, and diverse hardware environments.

%Current benchmarks formed by AutoDL mainly focus on a single constrained set of DL system components.  

%For example, NASBench-101~\cite{Chris2019ICML} that is a benchmark dataset for convolutional neural network architecture (i.e., CNN) would only include convolution system component (i.e., CONV$1 \times 1$, CONV$3\times3$,  and MAXPOOL$3\times3$) trained on CIFAR-10.

%a key question: \textbf{How can AutoML quickly and precisely recommend multi-dimensional system components without requiring any training feedback?} 

To this end, we propose a large-scale deep learning system component set comprising twenty-seven adjustable components. These components can be categorized into three dimensions: model and training components (including model architecture, training optimization, regularization and generalization, and framework), data components (e.g., size of training set, Input length, and Similarity between training and testing set), and hardware components (e.g., GPU type and number of GPUs). Each component includes a diverse array of candidate values, facilitating the configuration of DL systems across various model architectures, data augmentation techniques, and hardware setups. %Meanwhile, we have constructed a multi-dimensional dataset tailored for computer vision tasks

Based on the proposed system component dataset, we further introduce EfficientDL, an innovative deep learning board designed for automatic performance prediction and component recommendation. EfficientDL is powered by two key elements: a static performance prediction model and an optimized component recommendation algorithm, which enable EfficientDL to swiftly and precisely recommend twenty-seven system components and predict the performance of DL models without requiring any
training feedback. Specifically, we first present a static performance prediction model using Random Forest (i.e., RF) regression to eliminate reliance on actual running of parameterized models during the optimization search. Importantly, we enhance the component recommendation process through an improved Bayesian Optimization method, called $\alpha\beta$-BO search, to efficiently identify well-performing configurations within a reduced search space.

Experimental results on the CIFAR-10 dataset show that EfficientDL can rapidly recommend a deep learning system for image classification tasks, significantly outperforming state-of-the-art AutoML methods in both speed and accuracy (See in Figure~\ref{fig:performence}). Compared to existing state-of-the-art AutoDL frameworks (particularly Auto-PyTorch), our EfficientDL reduces the time by at least 20-fold while identifying a model with fewer parameters and superior performance, achieving a Top-1 accuracy of 91.31\%. Furthermore, the simplicity and power of EfficientDL stem from its compatibility with most DL models. For example, tests conducted on six popular model architectures, including ResNet50, MobileNetV3, EfficientNet-B0, MaxViT-T, Swin-B, and DaViT-T, demonstrate that EfficientDL achieves superior performance compared to handcrafted state-of-the-art DL systems. For instance, EfficientDL enhances Top-1 accuracy by 0.13\% to 0.69\%. Additionally, when employing the pre-trained models generated by our recommended components as backbones for object detection tasks, EfficientDL improves the mean Average Precision (i.e., mAP) by more than 0.49 for Swin-B and 0.22 for MobileNetV3.
We have released a replication package in the \texttt{GitHub repository \footnote{\url{https://github.com/OpenSELab/EfficientDL}}} to provide transparency into our research process.

In summary, the main contributions of our study are fourfold: 
\begin{enumerate}
     \item We propose a large DL system component set covering twenty-seven changeable components of DL, including model and training, data, and hardware dimensions.
     \item We construct ImageClassEval, a multi-dimensional, fine-grained dataset tailored for computer vision tasks. 
    \item We propose EfficientDL, an innovative deep-learning board designed for automatic performance prediction and component recommendation without relying on training feedback from the actual run in search optimization.

   % \item We design a static performance prediction method with a multi-dimensional dataset to eliminate the dependency on actual training and running current parameterized models in search optimization. 

%3) Experimental results on the CIFAR-10 dataset indicate that our automated performance estimation and recommendation framework, i.e., EfficientDL, can quickly recommend appropriate system components and exhibits promising compatibility across six well-defined model architectures.
 
\end{enumerate}

%% file: 03-relatedwork.tex
\section{Related work}\label{sec:related_works}

To assist developers in discovering suitable DL systems for their specific tasks, DL researchers primarily focus on automated DL techniques and well-designed benchmarks. Next, we will present the related work.

%have proposed different AutoDL methods, including neural architecture search, hyperparameter tuning, and benchmarks. 

\textbf{Automated Deep Learning (i.e., AutoDL) Techniques:} AutoDL techniques~\cite{8924184,gupta2020accelerator,lattuada2022performance,jin2023autokeras,liang2019evolutionary} refer to a suite of methods and tools designed to automate various stages of the deep learning pipeline, reducing the need for human intervention and expertise. Currently, these techniques tend to focus more on neural architecture search (i.e., NAS), and hyperparameter tuning. 

NAS~\cite{Thomas2019neural} automates the process of designing neural network architectures, which explores a predefined search space to find the optimal architecture for a given task with different search strategies (e.g., Reinforcement Learning (RL)~\cite{Barret2017ICLR, Barret2018cvpr}, Evolutionary Algorithms~\cite{Esteban2019AAAI}, and one-shot weight-sharing strategy~\cite{chen2021iccv}). For example, Tan et al.~\cite{tan2021efficientnetv2} optimize the combination of MBConv and Fuded-MBConv modules by NAS on EfficientNetV2, which improves the model efficiency and reduces the model size by 6.8 times. Gupta et al.~\cite{gupta2020accelerator} apply the NAS to search the convolution Block, convolution kernel, and activation function components. However, current NAS methods are mainly designed to search for a specific structure space, requiring users to predefine a search space as a starting point. 

Hyperparameter tuning involves finding the best set of hyperparameters (e.g., learning rate, batch size) that maximize the performance of a model architecture with grid search, random search, or Bayesian optimization~\cite{James2012JMLR}. For example, Diaz et al.~\cite{G2017IB} apply a derivative-free optimization tool to automatically and effectively search the appropriate parameters for neural networks. However, currently, most hyperparameter optimization methods are only treated as a subsequent step to NAS, ignoring the interaction between hyperparameters and choice of architecture. 

%Neural Architecture Search (i.e., NAS) is a subfield of automated deep learning that focuses on the automated design of neural network architectures to find the well-performing architecture for a given task, dataset, and set of constraints, without human intervention~\cite{Thomas2019neural}. Currently,  these methodologies tend to focus more on search strategies to explore the search space to boost the performance of deep learning models, including bayesian optimization, reinforcement learning, or evolutionary algorithms. 

\begin{figure}
    \centering
    \includegraphics[width=1.0\linewidth, scale=1.0]{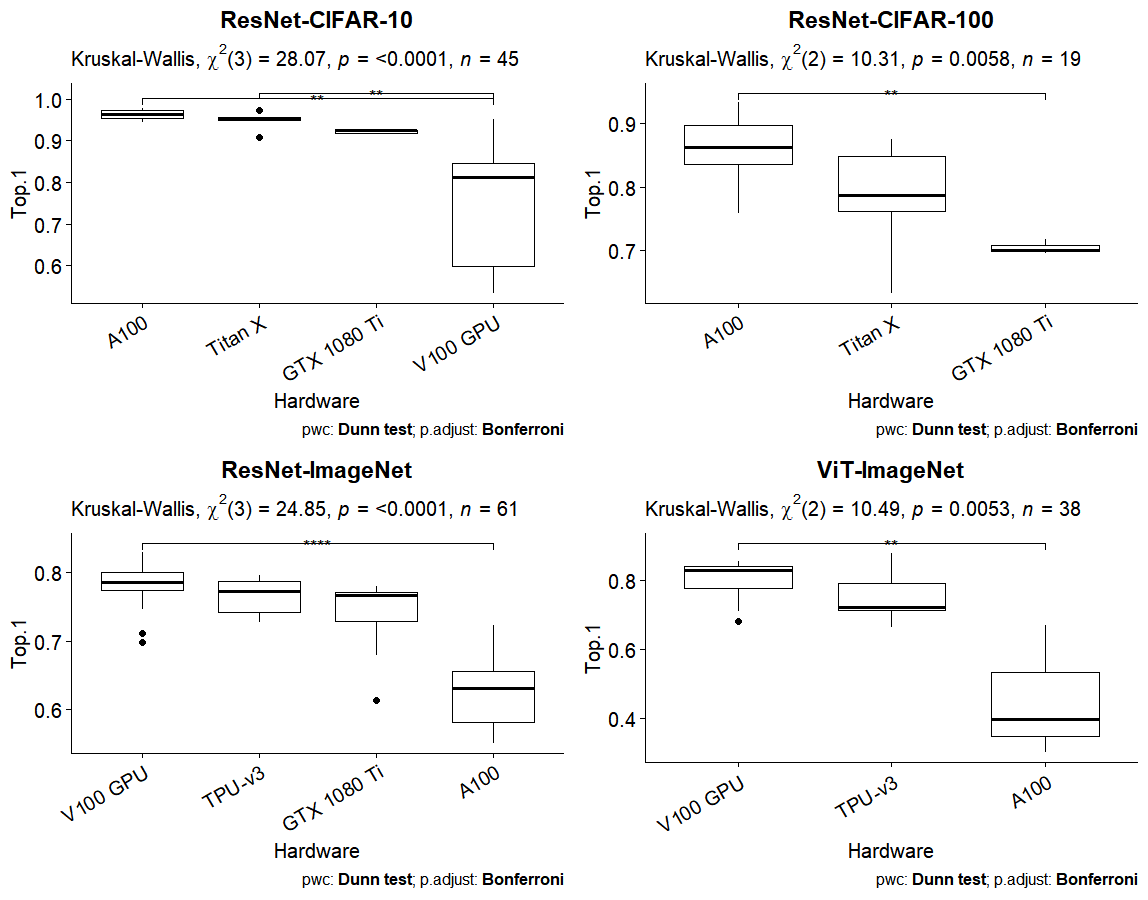}
    \caption{Top-1 accuracy of DL models with the same dataset run on different hardware. }
    \label{fig:hardware}
\end{figure}

AutoNet~\cite{Hector2016PMLR} is one of the first to jointly optimize architectures and hyperparameters at scale. Meanwhile, Lucas et al.~\cite{Lucas2021TPMI} improve the vague description~\cite{Hector2016PMLR} further to propose Auto-PyTorch. Although these methods have taken into account the interactions between model architecture and hyperparameters, they are also computationally intensive, requiring a lot of computing resources and time to explore the model space with the models’ training feedback under a given parameter. For example, Liang et al.~\cite{liang2019evolutionary} need tens of hours to use automated machine learning techniques for a dataset containing only 112,120 images. Furthermore, in addition to model architecture and hyperparameters, the performance of a deep learning system is also influenced by other factors (e.g., data distribution and hardware). As shown in Fig.~\ref{fig:hardware}, different hardware yields varying Top-1 accuracy for the same model and datasets on the image classification task. Therefore, our EfficientDL framework aims to quickly and precisely recommend twenty-seven system components and predict the performance of DL models without requiring any
training feedback.
%construct a comprehensive and fine-grained set of components that influence the performance of DL models to reach their full potential. 

%Facing this problem, our EfficientDL framework uses less time, provides a comprehensive set of deep learning components, and is able to give performance predictions directly without training feedback.

%\textbf{Hyperparameter tuning:} 

\textbf{Well-Designed Benchmarks:} Benchmarks for AutoDL are essential, which provide standardized datasets, tasks, and metrics to ensure a fair and consistent assessment of the performance of automated deep learning systems~\cite{Marius2020MLR}. For example, NASBench-101~\cite{Chris2019ICML} is a benchmark dataset for neural architecture search (NAS) that includes a fixed search space with 423,624 unique neural network architectures, each evaluated on CIFAR-10. Meanwhile, NASBench-101 provides the factors with architecture performance metrics, training time, and validation accuracy. NASBench-201~\cite{Xuan2020ICLR} is an extension of NASBench-101 with a different search space, covering three datasets: CIFAR-10, CIFAR-100, and ImageNet-16-120, which allows the study of NAS methods across multiple datasets and provides more comprehensive performance metrics. LCBench~\cite{Lucas2021TPMI} provides learning curves for various deep learning models trained with different hyperparameters on multiple datasets, which includes detailed information on the performance of models at different training epochs, enabling the analysis of training dynamics. HPOBench~\cite{Katharina2021NeurIPS} is a benchmark suite for hyperparameter optimization (HPO) that includes a collection of tasks from various domains such as image classification, NLP, and tabular data.

However, the current benchmarks mainly focus on limited factors (e.g., model architecture and hyperparameter) that are coarse-grained. A benchmark with appropriate fine-grained evaluation metrics that reflect the real-world performance of DL systems is essential for meaningful estimation and recommendation. To tackle the challenge of constrained search spaces, we propose a large DL system component set covering 27 changeable components of DL, which can be categorized into three dimensions: model and training dimension, data dimension, and Hardware dimension.

%% file: 05-DeepImageBenchdata.tex
\section{Multi-dimensional system component dataset}\label{sec:dataset}

% A comprehensive, multi-dimensional, fine-grained system component dataset
% is crucial for AutoML without requiring any training feedback during the traditional optimization search process. Due to the restricted nature of these search spaces, we propose a large-scale deep learning system
% component set comprising twenty-seven adjustable components. Additionally, based on the proposed system component set, we construct a multi-dimensional dataset tailored for computer vision tasks. 

\subsection{System component design}\label{sec:component_design}

% Based on the fact that the performance of DL models are sensitive to various factors, we define these various factors as the key components, which be categorized into model components (including architecture and hyperparameters), data components, and Hardware components DL models run on. Therefore, we systematically designed a comprehensive and fine-grained set of components that influence the performance of DL models.
The performance of DL models is influenced by a multitude of factors, which we regard as components of DL systems and categorize into three groups: model and training components (including architecture, training optimization, regularization and generalization, and framework), data components, and hardware components on which the DL models run. Based on this categorization, we design a comprehensive and fine-grained set of deep learning components.
%Based on the fact that performance of DL models is influenced not only by the model architecture and hyperparameters but also by the specific dataset and hardware environment, we systematically designed three key components, covering  model, data, and hardware.

\subsubsection{Model and training components}

Model and training components encompass a diverse set of modules, strategies, and parameter configurations that are essential for constructing, training, and optimizing deep learning models. These components work together to define the model’s architecture, learning capacity, generalization performance, and task-specific results. Broadly, they can be classified into four primary categories: model architecture, training optimization, regularization and generalization, and framework.

%refer to the components of a particular layer within the architecture of neural network and in particular to the hyperparameters related to that neural network, which can be categorized into two main types: model architecture components and hyperparameter components.

\textbf{Model architecture components} refer to the fundamental building blocks and overall design of the model. These components define how the model processes input data, how information is passed through the model, and how output is generated. Our study mainly focuses on Convolutional Neural Networks (e.g., ResNet~\cite{he2016deep}, MobileNetV2~\cite{sandler2018mobilenetv2}, and AlexNet~\cite{krizhevsky2017imagenet}) and Vision Transformers (e.g., Swin Transformer~\cite{liu2021swin} and DeiT~\cite{ding2022davit}), so the main model architecture components include:

\begin{itemize}[wide = 0pt, itemsep = 3pt]

\item \textbf{Normalization layer:} It helps address issues such as internal covariate shift, where the distribution of inputs to each layer changes during training, impeding network convergence. The list of normalization methods includes Layer Normalization, Batch Normalization, Spectral Normalization, Weight Normalization, etc.

%refers to a specialized neural network layer aimed to standardize the inputs to improve training stability and performance.
%refers to the component used to make optimization easier by smoothing the loss surface of the network. The list of normalization methods includes Layer Normalization, Batch Normalization, Spectral Normalization, Weight Normalization, etc.

\item \textbf{Initialization:} Initialization impacts deep learning performance by affecting gradient flow, optimization efficiency, and convergence speed, leading to better training and generalization. The list of initialization methods includes Kaiming Initialization, Xavier Initialization, Fixup Initialization, etc.

%refers to the component used to initialize the weights of neural networks. 

\item \textbf{Convolution:} It involves a learnable kernel sliding over the image and performing element-wise multiplication with the input. The specification allows for parameter sharing and translation invariance. The list of convolution operations includes Ordinary Convolution, Depthwise Convolution, Pointwise Convolution, Grouped Convolution, etc.

%is an operation that can be used to learn representations from images. 

\item \textbf{Skip connection: } Skip connections enhance model performance by improving gradient flow and information transfer, leading to better training stability and accuracy. The list of skip connections includes Residual connection, concatenated skip connection, Zero-padded shortcut connection, and Deactivable Skip connection, etc.

%refers to the component that enables layers to bypass intermediate layers and establish the connections with layers located further up the network, facilitating smoother information flow within the network

\item \textbf{Activation function:} Activation functions impact model performance by introducing non-linearity, enabling the network to learn complex patterns and improving its ability to make accurate predictions. The list of activation functions includes ReLU, GELU, Sigmoid Activation, Tanh Activation, Leak ReLU, etc.

%refers to the function that we apply in neural networks after (typically) applying an affine transformation combining weights and input features. The rectified linear unit, or ReLU, has been the most popular in the past decade, although the choice is architecture dependent and many alternatives have emerged in recent years. 

\item \textbf{Pooling operation:} Pooling operations are crucial for DL model performance as they extract key features, reduce spatial dimensions and computational complexity, improve robustness through translation invariance, and minimize overfitting by reducing parameters. The list of pooling operations includes Max Pooling, Average Pooling, Global Average pooling, Center Pooling, etc.

%It can also induce favourable properties such as translation invariance in image classification, as well as bring together information from different parts of a network in tasks like object detection (e.g. pooling different scales).
%refers to the component used to pool features together, often downsampling the feature map to a smaller size.

%\item \textbf{Data Augmentagion:} refers to the components that are designed to reduce the test error of a machine learning algorithm, possibly at the expense of training error. Many different forms of regularization exist in the field of deep learning. The list of regularization strategies includes Dropout, Label Smoothing, Attention Dropout, R1 Regularization, Stochastic Depth, L1 Regularization, etc.

\item \textbf{Feedforward network:} Feedforward networks are the foundation for many deep learning models, and their simplicity makes them easier to understand and implement compared to more complex networks like recurrent or convolutional neural networks. The list of feedforward networks includes Dense Connections, Linear Layer, Position-Wise Feed-Forward Layer, Feedforward Network, Adapter, etc.

%refers to  a type of neural network architecture which relies primarily on dense-like connections.

\item \textbf{Attention mechanism:} Attention mechanism assigns different levels of importance (weights) to different input elements based on their relevance to the current task, allowing the model to prioritize certain information over others. The list of attention mechanisms includes Scaled Dot-Product Attention, Multi-Head Attention, Fixed Factorized Attention, Additive Attention, etc.

%is a technique used in deep learning that enables models to selectively focus on specific parts of an input sequence when generating each element of the output sequence. 
%refers to a component used in neural networks to model long-range interaction, for example across a text in NLP. The key idea is to build shortcuts between a context vector and the input, to allow a model to attend to different parts. The list of Attention Mechanisms includes Scaled Dot-Product Attention, Multi-Head Attention, Fixed Factorized Attention, Additive Attention, etc.

\item \textbf{Output function:} Output function shapes the output format, optimization, gradient flow, and performance metrics, ensuring the model's suitability for various tasks. The list of output functions includes Softmax, Heatmap, Mixture of Logistic Distributions, Adaptive Softmax, PAFs, etc.

%refers to the layer used towards the end of a network to transform to the desired form for a loss function. For example, the Softmax relies on logits to construct a conditional probability.
\end{itemize}

\textbf{Training optimization components} refer to the various techniques, algorithms, and hyperparameters used to improve the performance, efficiency, and convergence of DL models during the training process. They are crucial for ensuring that the model learns effectively from the data, avoids issues like overfitting or underfitting, and converges to a solution in a reasonable amount of time. These components include:

\begin{itemize}[wide = 0pt, itemsep = 3pt]

\item \textbf{Learning rate schedule:} It controls the rate of learning over time, impacting convergence speed, training stability, and the model's ability to reach an optimal solution. The list of learning rate schedules includes Linear Warmup, Cosine Annealing, Step Decay, Linear Warmup with Decay, Linear Warmup with Cosine Annealing, etc.

%refers to the schedule for the learning rate during the training of neural networks.

\item \textbf{Optimization algorithm:} The optimization algorithm determines how these parameters are updated, thereby influencing the model's convergence speed and overall performance. The list of optimization algorithms includes Adam, SGD, Adafactor, RMSProp, AdaGrad, AdamW, SGD with Momentum, etc.

%refers to the process of adjusting model parameters (e.g., weights and biases) to minimize the loss function. 

\item \textbf{Size of parameter:} A higher number of parameters increases the model's expressive power but also adds computational complexity and the risk of overfitting.

%refers to the total number of trainable parameters (e.g., weights and biases) in the model. 

\item \textbf{Batch size:} Larger batch sizes provide more stable gradient estimates but require more computational resources; smaller batch sizes allow for more frequent updates, which may help escape local minima.

%refers to  the number of samples used in each parameter update iteration. 

\item \textbf{Learning rate:} determines the step size for updating the model's parameters in each iteration. An excessively high learning rate can lead to unstable training and non-convergence, while an overly low learning rate can result in slow convergence or getting stuck in local minima.

\item \textbf{Epochs: } More epochs enable the model to fit the data better but can also increase the risk of overfitting.
%More epochs allow the model to better fit the data but may also lead to overfitting.

%represent the number of times the entire training dataset passes through the model. 

\end{itemize}

\textbf{Regularization and generalization components} refer to techniques and strategies used in machine learning to improve a model's ability to generalize to unseen data and avoid overfitting during training. These components help the model perform well not only on the training data but also on new, previously unseen data. These components include:

\begin{itemize}[wide = 0pt, itemsep = 3pt]

\item \textbf{Regularization:} It helps prevent overfitting, improving the model’s ability to generalize to new, unseen data. The list of regularization strategies includes Dropout, Label Smoothing, Attention Dropout, R1 Regularization, Stochastic Depth, L1 Regularization, etc.

%refers to the component designed to reduce the test error of a machine learning algorithm, possibly at the expense of training error.

\item \textbf{Data augmentation:} Data augmentation would expand the dataset without incurring additional data collection costs, thereby improving the generalization ability of machine learning models, reducing the risk of overfitting, and enhancing the model's robustness in different scenarios. The data augmentation list includes image rotation, scaling, translation, flipping, cropping, color adjustment, noise addition, etc.

%refers to a technique used to increase the diversity of a training dataset by applying various transformations or operations to existing data to generate new instances.

\end{itemize}

\textbf{Framework} refers to a collection of tools and libraries designed to build, train, and deploy deep learning models. It provides a set of efficient and user-friendly programming interfaces and pre-built functional modules, simplifying the development process of deep learning models. The framework list includes PyTorch, Tensorflow, Caffe, Caffe2, etc.

\subsubsection{Data components}

Data components refer to the characteristics and structure of the dataset used to train, validate, and test DL models. These components are critical as they directly influence the models' performance, the design of the learning tasks, and how well the models generalize to new data. Key data components include:

%refer to the various data and processing methods used for the training and testing of DL models. These components are crucial for the model’s performance as they impact the model’s learning effectiveness, generalization ability, and inference accuracy. The data components can be further divided into the following key sub-components, each playing a specific role in the deep learning workflow.

\begin{itemize}[wide = 0pt, itemsep = 3pt]

\item \textbf{Size of training set:} refers to the total number of data instances used during model training. The size has a direct impact on the effectiveness of the model's training: a larger training set typically provides richer information, enhancing the model's learning capacity in handling new data. However, an excessively large training set may extend training duration and require more computational resources. Conversely, a training set that is too small cannot provide sufficient information, impeding the model's ability to generalize effectively.

\item \textbf{Size of testing set:} refers to the total number of data instances used to evaluate the model's performance after training, aimed at verifying the model's generalization ability and effectiveness in practical applications. The size of the testing set influences the stability and reliability of the evaluation results. 

\item \textbf{Input length:}  refers to the specific dimension or sequence length of each input sample that a deep learning model receives, determining the input shape and processing method of the model. The input length can vary depending on the type of data, such as, for image data, input length usually refers to the width and height of the image (in pixels) and the number of channels (e.g., 3 channels for an RGB image). For text data, input length usually refers to the number of words or characters in a text sequence.

\item \textbf{Output length: } refers to the specific dimension or sequence length of the output generated by a deep learning model, determining the model's output shape and designed according to the task's requirements. The output length depends on the type of model and the application scenario, closely aligning with the model's intended task. For instance, for classification tasks, the output length typically equals the number of classification categories. 
For regression tasks, the output length is usually 1, representing the prediction of a single continuous value. However, in multi-dimensional regression, the output length may be greater than 1.

\item \textbf{Similarity between training and testing set: } refers to the degree of resemblance between the training data and the test data in terms of feature space or distribution. High similarity usually indicates that the training set and test set come from similar distributions, which can lead to better model performance on the test set because the inputs it encounters during testing are similar to those seen during training. In our study, cosine similarity~\cite{Xia2015ins}, Jensen-Shannon (JS)~\cite{Menendez1997JFI}, and L2 distance~\cite{Ruschendorf1990JMA} serve as the metrics for assessing the similarity between training and testing sets. 
\end{itemize}

\subsubsection{Hardware components}

Hardware components refer to the physical computing resources and devices that perform model training and inference tasks. The performance and configuration of the hardware have a direct impact on the speed, efficiency, scalability, and overall effectiveness of deep learning models. In our study, we would focus on the GPU devices, which have a significant impact on the performance of deep learning models, primarily in accelerating the training process, reducing training time, and improving model scalability.

\begin{itemize}[wide = 0pt, itemsep = 3pt]

\item \textbf{Type of GPU:} refers to the different kinds of GPUs used for handling deep learning tasks. Various types of GPUs offer different computational power, memory size, power consumption, and architectural design, all of which determine their performance in deep learning models. Selecting the appropriate GPU type is crucial for enhancing model training and inference efficiency. The type of GPU list includes V100, GeForce GTX 1080 Ti, NVIDIA RTX 2080 Ti, NVIDIA GTX980, etc.

\item \textbf{Number of GPUs:} refers to the number of GPUs used to execute deep learning tasks. The number of GPUs directly impacts the training speed, parallel computing capability, and overall performance of a deep learning model. Proper configuration of the GPU quantity can significantly enhance model training efficiency, reduce training time, and enable large-scale deep learning tasks to be completed within an acceptable timeframe.
\end{itemize}

The configuration space for each fine-grained component is presented in Table \ref{tab:model_architecture} of Appendix. From Table \ref{tab:model_architecture}, we can observe that most of model architecture components and Hardware components are discrete values, while data components and training optimization components are continuous values. Furthermore, not all components exhibit unique values, as they may represent a combination of several discrete values. For example, Pooling Operations used in ResNet~\cite{he2015axiv} include Average Pooling, Global Average Pooling, and Max Pooling.

\subsection{Multi-dimensional dataset construction}\label{sec:multi-dataset}

Drawing upon the designed component set, our objective is to construct a multi-dimensional dataset to facilitate the development of the static performance prediction method, aiming to eliminate the dependency on actual training and running current parameterized models in search optimization. Given that the designed metrics encompass various model structures, parameters, and GPU types, the cost of conducting these experiments is substantial. As a result, we focus on gathering multi-dimensional datasets from existing published research papers that have been generally recognized to be the top peer-reviewed and influential publication. Specifically, the construction of  multi-dimensional dataset is mainly divided into four steps: %(1) Collecting papers for the specific task with DL models;(2) Filtering out models that include code and parameters; (3) Extracting values of metrics and performance for each of DL models; (4) Encoding the metrics to form multi-dimensional dataset for the specific task.

\smallskip\noindent\textbf{Searching for primary studies on specific tasks using DL models.} We first search google scholar\footnote{https://scholar.google.ca/} for primary studies concentrating on the application of DL models in addressing specific tasks (e.g., image classification, text classification) with the terms of specific tasks or deep learning. We then go through the abstract of all the searched papers to focus on the specific task with the DL models. Subsequently, we perform a forward and reverse snowball search~\cite{Wohlin2012eise}, tracing all the papers cited by each selected study. We carefully review these references to identify additional relevant studies. This process is repeated recursively for each newly collected paper until all pertinent studies are exhausted.  

\smallskip\noindent\textbf{Excluding studies without code and parameters.} 
% Since we focus on a relatively comprehensive and fine-grained component set that should be extracted from code and parameters of pre-trained models, primary studies without code and parameters of pre-trained models should be excluded. Therefore, 
We manually review each searched primary study by step 1 to find whether this primary study includes the code and parameters of pre-trained models. We then only retain these primary studies that include both code and parameters to construct the multi-dimensional dataset.

\smallskip\noindent\textbf{Extracting values of fine-grained components and performance for each DL model included in primary studies.} We primarily use a combination of automated and manual methods to extract values of fine-grained components and performance for each DL model. Meanwhile, it is important to note that a primary study would include a specific DL model with different values of components and different compared methods, which are treated as multiple instances in our study. Specifically, we first apply regular expressions to automatically identify keywords and corresponding table information (e.g., model parameters, hyperparameters, and performance) in the primary studies to extract values of components. In case where certain components are not explicitly described in the paper, we check the code of DL models to locate the corresponding values. Finally, to ensure the accuracy of the extracted values, we conduct a manual examination of the relevant papers and code, supplementing any missing component values as necessary. Additionally, components that are neither documented in the papers nor the code are recorded as missing values.

\smallskip\noindent\textbf{Encoding fine-grained component to form multi-dimensional dataset for the specific task.} As presented in Section~\ref{sec:component_design}, many components, particularly those related to model architecture, are composed of discrete values, with some being different combinations of multiple discrete values. For example, data augmentation component can be a combination of Random Horizontal Flip, Random Resized Crop, and Mixup. Therefore, we apply distinct encoding methods (i.e.,  one-hot encoding, label encoding, and polynomial encoding) to transform discrete values into a numerical form based on the nature of components. Specifically,

\begin{itemize}[wide = 0pt, itemsep = 3pt]

\item For components with only two non-parallel discrete values, we apply one-hot encoding method to encode, such as Skip Connections, Output Functions.

\item For components with multiple discrete values that are not parallel, we apply label encoding method to encode, such as Initialization, Position Embeddings, Attention Mechanisms, Learning Rate Schedules, and etc.

\item For components with multiple parallel discrete values, we apply polynomial encoding method to encode, such as Normalization, Convolutions, Activation Functions, Pooling Operations, and etc.

\end{itemize}

\subsection{ImageClassEval: A dataset for performance evaluation and component recommendation of DL models on image tasks}\label{sec:ImageData}

Followed by the process of constructing a multi-dimensional dataset in Section~\ref{sec:multi-dataset}, we focus on image classification tasks and develop the ImageClassEval dataset. Next, we provide a detailed overview of the ImageClassEval dataset construction, highlighting its properties, including scale, hierarchy, and diversity.

%To validate the effectiveness of our EfficientDL framework, we focus on image classification tasks and develop the ImageClassEval dataset. 

%ImageClassEval is specifically designed to support automated performance evaluation and component recommendation for image classification models. Next, we provide a detailed overview of the ImageClassEval dataset construction, highlighting its properties, including scale, hierarchy, and diversity.

\subsubsection{Constructing ImageClassEval}

ImageClassEval is a comprehensive, multi-dimensional, fine-grained system
component dataset for image tasks, which enables researchers and practitioners to develop a static performance prediction model and optimized component
recommendation algorithm in a data-driven manner, removing the dependency on actually running parameterized models during the traditional optimization search process. We describe here the process of constructing the ImageClassEval dataset:

%with multi-dimensional dataset construction method in Section~\ref{sec:multi-dataset}. Specifically,

%dataset that enables us to develop a static performance prediction model and comprehensive optimized component recommendation algorithm

%for performance evaluation and component recommendation of DL models on image tasks, 

%which can be used to evaluate the effectiveness of our EfficientDL framework. 

\smallskip\noindent\textbf{Collecting and cleaning candidate primary studies on image classification tasks with DL models.} We first collect 1232 primary studies related to image classification tasks with different deep learning models from 2016 to 2023. We then clean candidate primary studies to gather a total of 342 candidate primary studies with code and parameters of pre-trained models. 

\smallskip\noindent\textbf{Extracting values of fine-grained components and performance for each DL model.} We construct the final ImageClassEval dataset by extracting values of fine-grained components and performance for each DL model. Meanwhile, in our ImageClassEval dataset, we use Top-1 accuracy as the metric for evaluating DL models, since most of the candidate primary studies report this metric. 

Through automatically identifying keywords and tables in papers and code, we gather a total of 2,433 DL models. Subsequently, after a manual review process, 741 DL models lacking fine-grained components or performance metrics are excluded. Finally, we construct the ImageClassEval dataset with 1,655 DL models.

\subsubsection{Properties of ImageClassEval}

ImageClassEval exhibits several key attributes that significantly enhance its utility and effectiveness for research and application, which can serve as a foundational resource, offering a comprehensive and well-structured collection of data points that enable researchers and practitioners to better predict the performance and optimize deep learning models across a wide array of configurations. We briefly explain below the various key attributes that our ImageClassEval exhibits.

\begin{figure}
    \centering
    \includegraphics[width=1.0\linewidth, scale=1.0]{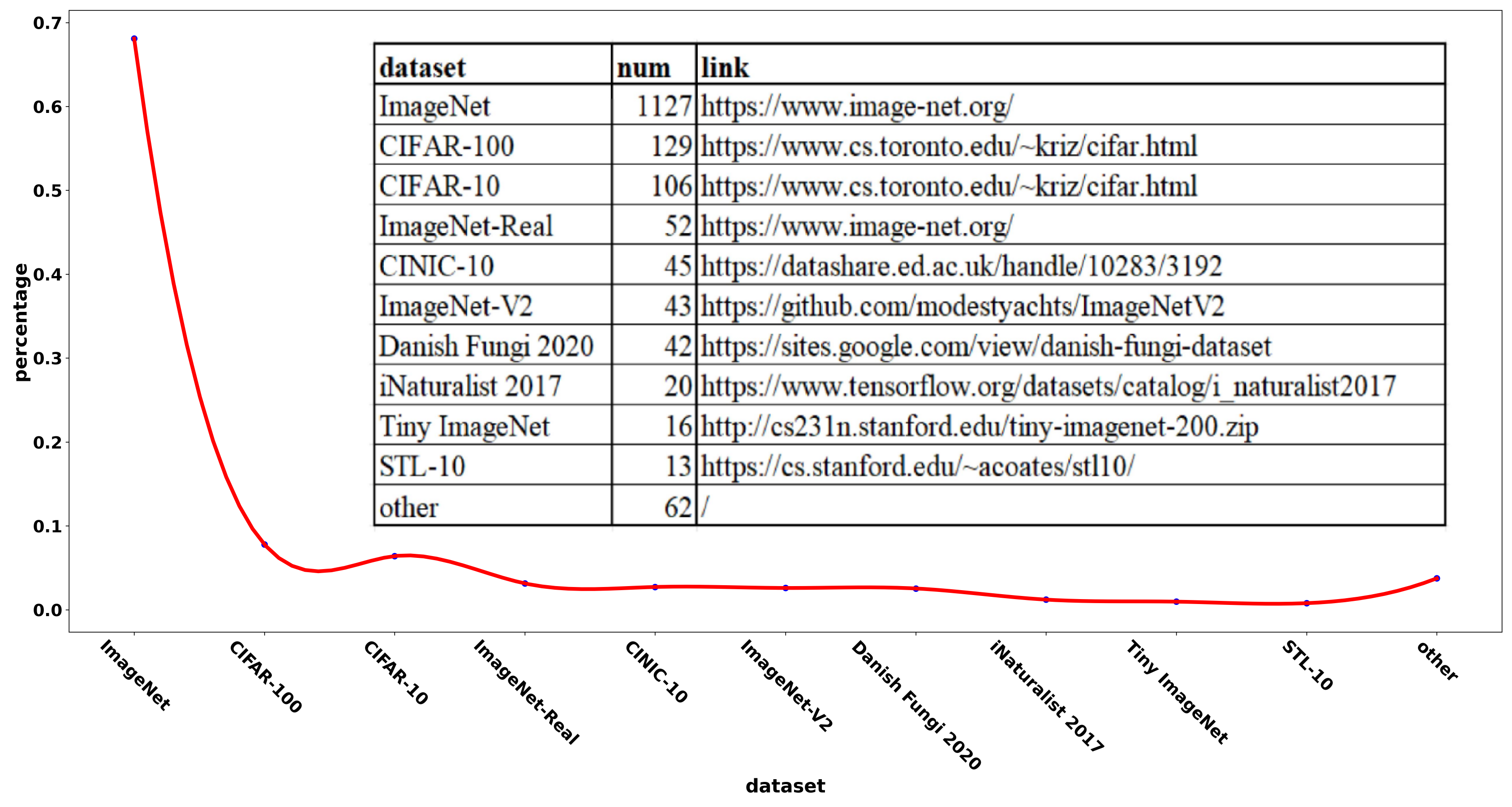}
    \caption{Distribution of these datasets utilized for training deep learning models. }
    \label{fig:data_distribution}
\end{figure}

\smallskip\noindent\textbf{Scale.} ImageClassEval is designed to offer the most comprehensive and diverse representation of image datasets available. At present, ImageClassEval encompasses 30 distinct image datasets, with Figure~\ref{fig:data_distribution} illustrating the distribution of these datasets utilized for training deep learning models. To the best of our knowledge, ImageClassEval constitutes the largest dataset for performance estimation and component recommendation of deep learning models, specifically tailored for the vision and AutoML research communities. This scale is reflected not only in the sheer number of image datasets but also in the inclusion of highly detailed, fine-grained components.

\begin{figure}
    \centering
    \includegraphics[width=1.0\linewidth, scale=1.0]{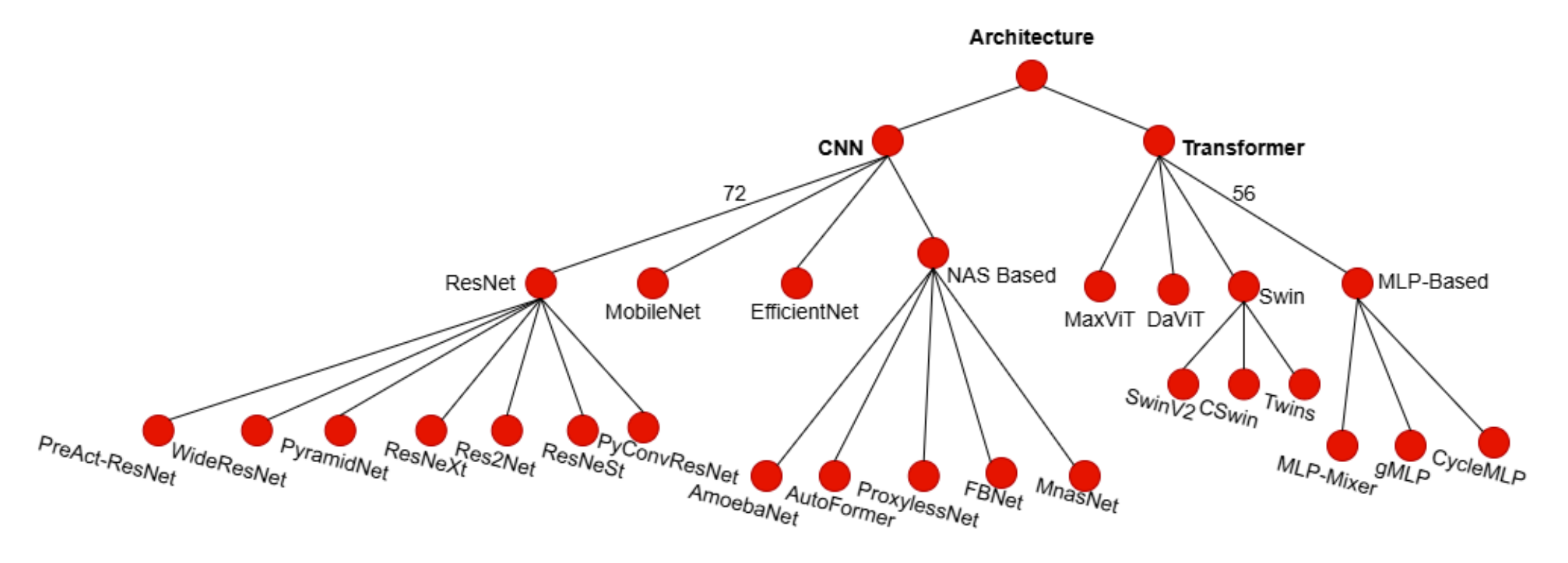}
    \caption{Hierarchical tree of model architecture in our ImageClassEval. }
    \label{fig:model_architecture}
\end{figure}

\smallskip\noindent\textbf{Hierarchy.} ImageClassEval is meticulously organized to reflect the hierarchical nature of deep learning model architectures, with a particular focus on Convolutional Neural Networks (i.e., CNNs) and Transformer-based models. Figure~\ref{fig:model_architecture} illustrates the hierarchical tree of architecture. 

For CNNs, the dataset captures the hierarchical design inherent to these models, encompassing mid-level architectural variations, such as the incorporation of residual connections (as seen in ResNet), the depth of the network. Meanwhile, it includes lower-level details such as the types of pooling operations (e.g., max-pooling or average-pooling) and the application of techniques like batch normalization in Table~\ref{tab:model_architecture} of Appendix. In the context of Transformer-based architectures, ImageClassEval documents variations across models such as MaxViT, DaViT, and Swin Transformer. This includes differences in layer depth, positional encoding strategies, and the size of feedforward networks.

This hierarchical organization not only facilitates a granular analysis of how different architectural components influence overall model performance but also enables meaningful comparisons across different deep learning paradigms. By providing a structured framework, ImageClassEval supports both detailed technical assessments and broader evaluations between various model types, thereby aiding in the development of more effective deep learning system recommendations.

\smallskip\noindent\textbf{Diversity.} ImageClassEval encompasses a rich variety of system components, including a broad spectrum of network architectures, hyperparameters, data components, and hardware components. For example, Table~\ref{tab:model_architecture} presents  our ImageClassEval with other datasets, including NASBench-101, NASBench-201, LCBench, and HPOBench. From Table~\ref{tab:model_architecture}  of Appendix, ImageClassEval includes 27 system components, while NASBench-101 only includes 2 system components (i.e., Convolutions and Pooling operations), NASBench-201 includes 3 system components, LCBench and HPOBench include 4 system components. 

Meanwhile, this diversity is not only reflected in the sheer number of configurations but also in the extensive range of options within each component. By offering a comprehensive selection of classical and cutting-edge models, as well as modern and traditional optimization methods, ImageClassEval provides researchers with the flexibility to explore and evaluate deep learning models across a vast landscape of possible scenarios. This contrasts with other datasets like NASBench-101 and NASBench-201, which are primarily focused on neural architecture search, or LCBench and HPOBench, which are more narrowly tailored towards hyperparameter optimization. 

Therefore, the diversity of ImageClassEval makes it an invaluable resource for advancing research in deep learning by enabling more robust and comprehensive performance evaluations across a broader set of configurations.

%下面的这个要放到实验部分去

%% file: 04-EfficientDL.tex
\section{EfficientDL: an innovative deep learning board}\label{sec:efficientDL}

\begin{figure*}
    \centering
    \includegraphics[width=0.95\linewidth, scale=0.8]{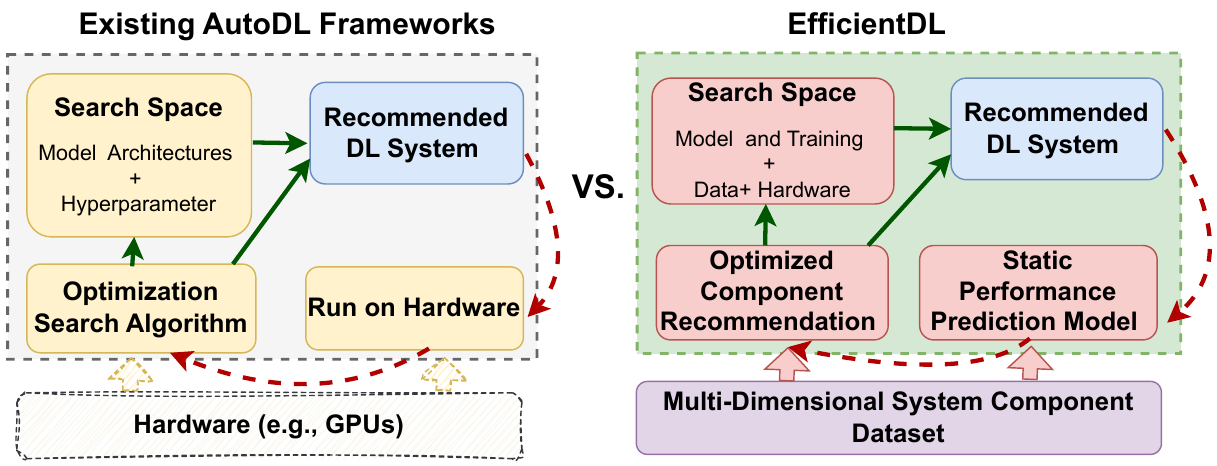}
    \caption{Comparison of AutoML frameworks. Existing AutoDL frameworks rely on training feedback from actual runs on the hardware. In contrast, our EfficientDL can quickly and precisely recommend twenty-seven system components and predict the performance of DL models without requiring any
training feedback, due to our multi-dimensional system component dataset. }
    \label{fig:overview}
\end{figure*}

% Current automatic DL frameworks depend on training feedback from actual runs with specific parameters to select suitable DL systems. To overcome this limitation,
In this section, we introduce EfficientDL, an innovative deep learning platform built on the comprehensive, multi-dimensional, fine-grained system component dataset outlined in Section~\ref{sec:component_design}. EfficientDL is designed for automatic performance prediction and component recommendation without relying on training feedback. As shown in Fig. ~\ref{fig:overview}, EfficientDL includes a static performance prediction model and a comprehensive optimized component recommendation algorithm. The static performance prediction model removes the need for training feedback from actual runs during the optimization process, while the optimized recommendation algorithm efficiently identifies well-performing configurations within a reduced search space.

%Fig.~\ref{fig:overview} presents an overview of EfficientDL’s architecture, including a static performance prediction model and a comprehensive optimized component recommendation algorithm. The static performance prediction model would help to eliminate the dependency on training feedback from actual runs under given parameters during the optimization search. The comprehensive optimized component recommendation algorithm would help to efficiently identify well-performing configurations within a reduced search space.

%We now present EfficientDL, a standardized pipeline for automating efficient performance prediction and component recommendation, including system component design, multi-dimensional dataset construction, static performance evaluation, component recommendation confirmation, and well-performing configurations recommendation. Within EfficientDL, the procedure for system component design involves constructing comprehensive and fine-grained components to expand search spaces and facilitate interactions among components. The utilization of multi-dimensional dataset and static performance evaluation can help design a static performance prediction method to eliminate the dependency on actual training and running current parameterized models in search
%optimization. Important component analysis and recommendation focus on components that are relatively important for the task to save search space and time. Fig.~\ref{fig:overview} presents an overview of EfficientDL’s architecture.

\subsection{Static performance prediction model}\label{sec:static_model}

Existing AutoDL methods rely on actual runs to assess the effectiveness of recommended DL systems, significantly increasing the time costs associated with identifying the optimal system components. 
Fortunately, our constructed dataset outlined in Section \ref{sec:dataset} introduces a paradigm shift. It enables the development of a static performance prediction model in a data-driven manner, capable of directly predicting model performance without the need for actual training iterations.
By leveraging this advanced framework, we can avoid the time-consuming training processes inherent in traditional methods and achieve an immediate evaluation of the recommended systems' effectiveness, substantially accelerating existing AutoDL processes in identifying the optimal system components. 

Specifically, we employ the random forest model \footnote{\url{https://scikit-learn.org/stable/modules/generated/sklearn.ensemble.RandomForestRegressor.html}} as the performance prediction model and train it on our multi-dimensional system component dataset. The diversity and comprehensiveness of the proposed dataset maximize the reliability of the performance prediction model. Besides, to further optimize the performance of this prediction model, we use a grid search strategy \footnote{\url{https://scikit-learn.org/stable/modules/generated/sklearn.model_selection.GridSearchCV.html}} to tune the hyperparameters of Random Forest learner, with the range of ``n\_estimators" being $\{5, 10, 30, 50, 80, 100, 150, 180, 200, 250, 280, 300\}$ and the range of ``max\_depth" being $\{3, 5, 10, 15\}$. 
The choice of Random Forest for building a performance prediction model on deep learning model evaluations is motivated by its reduced susceptibility to overfitting and efficient handling of high-dimensional data. Meanwhile, comparing with neural network, Random Forest can provide feature importance scores, helping to interpret which features most significantly impact the model’s predictions.

\subsection{Comprehensive optimized component recommendation algorithm}

% As outlined in Section~\ref{sec:component_design}, the performance of DL systems would be impacted by various system components. 
Our comprehensive optimized component recommendation algorithm aims to efficiently identify optimal component configurations. This advanced algorithm is powered by two key elements: component recommendation confirmation and Well-performing configurations recommendation.

\subsubsection{Component Recommendation Confirmation}\label{sec:component_recommendation}
As outlined in Section~\ref{sec:component_design}, the performance of DL systems would be impacted by various system components. 
While our proposed static performance prediction model substantially mitigates the time costs in performance evaluation, the simultaneous optimization of all twenty-seven components introduces an expansive search space, potentially increasing the computational burden of the optimization algorithm.
Moreover, the complete component recommendation may be unnecessary in certain scenarios where 
specific components are predetermined due to constraints or user preferences.  For instance, when the system is restricted to an NVIDIA RTX 2080 Ti GPU, there is no need to recommend the GPU type.
To address these considerations, our EfficientDL provides two strategies for component recommendation confirmation: one strategy involves manually specifying the components and their applicable ranges, while the other strategy automatically identifies the necessary components through an analysis of their importance.   
To automatically identify the necessary recommended component, we apply permutation feature important~\cite{Andre2010Bio} to evaluate the importance of components on our static performance prediction model and recommend the Top-5 important components. For the range of each recommended component, we use the searched space outlined in Table~\ref{tab:model_architecture} of the Appendix.

\begin{figure}
    \centering
    \includegraphics[width=0.95\linewidth, scale=1.0]{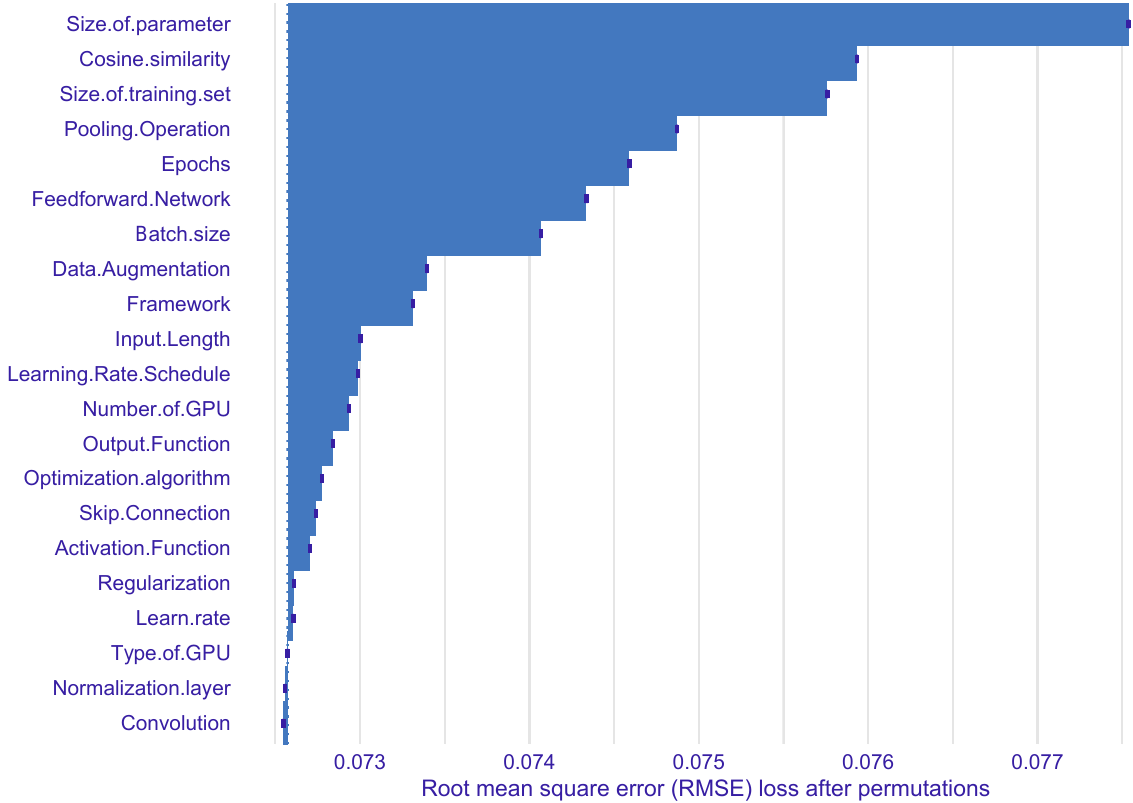}
    \caption{Components Recommendation Confirmation on the ImageClassEval dataset for image-related tasks. }
    \label{fig:feature_important}
\end{figure}

Figure~\ref{fig:feature_important} shows the recommended components for image-related tasks based on our ImageClassEval dataset.
From Figure~\ref{fig:feature_important}, we can observe that in image-related tasks, some model architecture components (i.e., Size of parameter, Pooling Operations, Feedforward Networks, Data Augmentation, and Framework),  some hyperparameter components (i.e., epochs and Batch size ) and some data components (i.e., cosine similarity between training and testing set, Size of training set, and Input Length) are the Top-10 key components. Therefore, researchers and practitioners in the field of image-related domain could pay close attention to the parameter settings of these components when addressing their specific challenges.

\subsubsection{Well-performing configurations recommendation for DL system components}
After confirming the key components required for optimization in Section~\ref{sec:component_recommendation}, our EfficientDL employs an improved Bayesian optimization strategy, namely  $\alpha\beta$-BO search, to recommend well-performing configurations for these essential DL system components. 
Bayesian optimization~\cite{Jonas1994JGO} enables the search process to rapidly converge to an optimal solution by dynamically balancing the exploration of untested regions with the exploitation of already identified promising areas. 
Typically, it is an iterative method that can efficiently explore and optimize objective functions by constructing a surrogate model, including three steps: building the surrogate model, selecting the next sampling point, and updating the surrogate model iteratively. Specifically, it begins by constructing a probabilistic model, typically a Gaussian Process, based on the existing sample data to approximate the distribution of the objective function. Then, an acquisition function is used to assess the sampling value of different points, balancing the exploration of new areas with the exploitation of known promising regions. In each iteration, the acquisition function evaluates potential sampling points and selects the next one, typically in the region most likely to improve the objective function. Once the sampling point is selected, the objective function's value is evaluated, and the surrogate model is updated, gradually narrowing the search for the optimal solution. This process is iterated until convergence to the optimal solution of the objective function or until the predefined evaluation limit is reached.

However, within the current Bayesian optimization framework, the use of Expected Improvement (EI) or Probability of Improvement (PI) as acquisition functions often prioritizes points with the highest expected values or the highest probabilities of improvement. While this strategy can be effective, it may inadvertently overlook regions of the search space with significant potential, such as points with lower expected values but higher probabilities, or those with higher expected values but lower probabilities. This limitation becomes especially pronounced when dealing with complex or multimodal objective functions, where the optimization process is at risk of becoming trapped in local optima, thereby failing to thoroughly explore the entire search space.

To address these limitations, we propose $\alpha\beta$-BO search, an advanced optimization algorithm designed to explore the search space more effectively using a novel acquisition function and exploration strategy. The $\alpha\beta$-BO Search method is conceptually straightforward and is implemented based on the Expected Improvement (EI) acquisition function, which can be seamlessly integrated with most surrogate models. This method enhances the exploration of previously uncharted regions by focusing on areas with a higher probability of yielding optimal values that have not been explored yet, while still building upon the information from previously explored areas. Additionally, to mitigate the risk of local convergence, $\alpha\beta$-BO Search incorporates a probability parameter, $\Omega$, which facilitates random searches throughout the exploration cycle. 

The formula of probability parameter ($\Omega$) for random search is as: 
\begin{equation}\label{eq:Om}
\Omega = 1 - k \times P
\end{equation}

In Eq.~\ref{eq:Om},  $k$ represents the number of repeated random explorations, and if one exploration is not random, $k$ is reduced to 1. $P$ denotes the user-defined likelihood of random exploration, where a higher value corresponds to a lower probability of random exploration. Through Eq.~\ref{eq:Om}, we can observe that over time, $\Omega$ decreases, refining the search focus. 

Meanwhile, the $\gamma$EI represents an improved algorithm for the EI acquisition function in our $\alpha\beta$-BO search. The formula of the $\gamma$EI acquisition function is as: 
\begin{equation}\label{eq:I}
\begin{split}
\gamma \text{EI}(\mathbf{x}) = \alpha \cdot \int_{-\infty}^{(f_{\text{best}} - m(\mathbf{x})) / \sigma(\mathbf{x})} (f_{\text{best}} - f(\mathbf{x})) \\ \cdot p(f(\mathbf{x}) | \mathbf{x}, D) \, df(\mathbf{x}) + \beta \cdot P(f(\mathbf{x}) > f_{\text{best}})
\end{split}
\end{equation}

In Eq.~\ref{eq:I}, \( f(\mathbf{x}) \) be a multi-dimensional Gaussian process as:
\begin{equation}
f(\mathbf{x}) \sim \mathcal{GP}(m(\mathbf{x}), k(\mathbf{x}, \mathbf{x}')) 
\end{equation}

\( m(\mathbf{x}) \) is the mean function, and \( k(\mathbf{x}, \mathbf{x}') \) is the covariance function. Additionally, in equation~\ref{eq:I}, \( f_{\text{best}} \) is the current best observed target value, \( \sigma(\mathbf{x}) \) is the standard deviation of the Gaussian surrogate model at position \( \mathbf{x} \), and \( p(f(\mathbf{x}) | \mathbf{x}, D) \) is the probability distribution of the target value at position \( \mathbf{x} \) given the observed data \( D \).

Indeed,  within a multi-dimensional Gaussian surrogate model, the derivation of the $\gamma$EI acquisition function is inherently complex, as it involves the cumulative distribution function (CDF) and probability density function (PDF) of the multivariate normal distribution.  Therefore, we define a vector \( \mathbf{Z} \) as:
\begin{equation}
\mathbf{Z} = \frac{(f_{\text{best}} - m(\mathbf{x}))}{\sigma(\mathbf{x})}.
\end{equation}

Then, the formula of the $\gamma$EI acquisition function can be further simplified to:
\begin{equation}\label{eq:ge}
\begin{split}
\gamma EI(\mathbf{x}) = \alpha \cdot (\sigma(\mathbf{x}) \cdot \mathbf{Z} + (f_{\text{best}} - m(\mathbf{x})) \cdot \Phi(\mathbf{Z})) \\ + \beta \cdot ( P\left(Z > \frac{f_{\text{best}} - m(\mathbf{x})}{\sigma(\mathbf{x})}\right)).
\end{split}
\end{equation}

In Eq.~\ref{eq:ge}, \( \Phi(\mathbf{Z}) \) is the cumulative distribution function (CDF) of the multivariate normal distribution. By applying the CDF of the standard normal distribution, this expression can be further transformed into the following form:
\begin{equation}
\begin{split}
\gamma EI(\mathbf{x}) = \alpha \cdot (\sigma(\mathbf{x}) \cdot \mathbf{Z} + (f_{\text{best}} - m(\mathbf{x})) \cdot \Phi(\mathbf{Z})) \\ + \beta \cdot ( 1 - \Phi\left(\frac{f_{\text{best}} - m(\mathbf{x})}{\sigma(\mathbf{x})}\right)).
\end{split}
\end{equation}

The detailed process of our $\alpha\beta$-BO search method is shown in the Algorithm~\ref{alg:bo_algorithm} of Appendix B.

%% file: 06-Experiment.tex
\section{Experiment}\label{sec:Experiment}

In this section, we first conduct experiments to evaluate the effectiveness of our EfficientDL in recommending DL models for image classification tasks, comparing their performance against the state-of-the-art AutoDL tools. We then do experiments to evaluate the effectiveness of the important components recommended for the existing DL models for image classification tasks. Furthermore, we conduct ablation studies to thoroughly analyze the significance of various components within our EfficientDL framework, with a particular focus on the effectiveness of the system components we extracted, the key component recommendations, and the optimization strategies applied. Finally, we apply the retrained backbone networks to downstream object detection tasks, providing evidence of the comprehensive effectiveness of our approach.

\subsection{Comparison with state-of-the-art AutoDL tools}

% We primarily evaluate the benefit of our EfficientDL framework in terms of its network architecture search capabilities, including baselines, implementation details and experimental analysis.

\subsubsection{Baselines for model architecture recommendation}

We compare EfficientDL to several state-of-the-art AutoDL frameworks focusing on neural architecture search, which are elaborated as follows:

\smallskip\noindent\textbf{Auto-Pytorch} ~\cite{Lucas2021TPMI} is an AutoDL framework utilizing multi-fidelity optimization to jointly optimize deep neural network architectural parameters (i.e., number layers, and max. number of units).

\smallskip\noindent\textbf{Auto-Keras} ~\cite{haifeng2023jml} is an efficient neural architecture search framework, which utilized Bayesian optimization to guide through the search space by selecting the most promising operations each time. Auto-Keras can search for a specific type of neural network architecture, e.g., multi-layer perceptron and convolutional neural network. For the convolutional neural network, Auto-keras uses the three-layer convolutional neural network as the default architecture and optimize the components including kernel\_size, number\_blocks, number\_layers, filters, dropout, and max\_pooling.

\smallskip\noindent\textbf{NASH} ~\cite{Thomas2017axiv} apply a simple hill-climbing procedure to search for convolutional neural network automatically. Meanwhile, NASH uses the three-layer convolutional neural network as the default architecture and optimizes the components including kernel\_size, number\_blocks, Skip Connections, and number\_layers.

\smallskip\noindent\textbf{NASBOT}~\cite{Kirthevasan2019axiv} is a Gaussian process-based Bayesian Optimisation framework for neural architecture search. Similarly, NASBOT also focuses on searching for convolutional neural networks automatically by optimizing the components including kernel\_size, number\_blocks, Skip Connections, dropout, and number\_layers.

\subsubsection{Implementation Details}

Similar to prior studies ~\cite{haifeng2023jml}~\cite{Lucas2021TPMI}, the comparative experiments are conducted on the CIFAR-10 dataset for image classification tasks to fairly  evaluate the benefits of our EfficientDL framework. Specifically, we also divide the original training data of CIFAR-10 into training and validation data by 8:2. We run each method on a single GPU (Nvidia GeForce RTX 3090) for 200 epochs with a batch size of 64. Consistent data augmentation techniques (i.e., horizontal flipping, random crops, and mixup) are employed across all methods. Furthermore, for a fair comparison, our EfficientDL framework focuses on the model architecture components including Convolutions, Skip Connections, Regularization, Activation Functions, Initialization, Pooling Operations, Normalization, and size of parameters. We report the Top-1 accuracy on the test set for all experiments.

\subsubsection{ Experimental Analysis.}

\smallskip\noindent\textbf{Compared to the frameworks under evaluation, our EfficientDL demonstrates the highest Top-1 accuracy while maintaining a relatively compact size of model architecture.} Table~\ref{table: nas} shows the size of parameters included in the neural architecture searched by compared methods and the Top-1 accuracy on the test dataset of CIFAR-10 with the best model architecture. From Table~\ref{table: nas}, we can observe that the model architecture recommended by our EfficientDL achieves a Top-1 accuracy of 91.31\% on CIFAR-10 after 200 epochs of training, which is a 1.31\% improvement over Auto-PyTorch, while also reducing the model size by 1.7M parameters. 

More importantly, our EfficientDL can search a relatively optimal model architecture in just 0.5 hours, whereas other methods require at least 10 hours. This highlights the effectiveness of our framework's static performance prediction model, which effectively eliminates reliance on exhaustive training and running of DL systems under recommended parameters during search optimization.

\begin{table}[t]
\centering
\caption{Results of neural architecture searched by compared methods on CIFAR-10.}\label{table: nas}
%\hspace{0.5cm}
%\setlength\tabcolsep{2pt}
\newcommand{\tabincell}[2]{\begin{tabular}{@{}#1@{}}#2\end{tabular}}
\centering
\begin{tabular}{llll} 
\hline
\textbf{Methods} & \textbf{Top-1 accuracy}& \textbf{Size of parameters} & \textbf{Hours} \\
\hline
NASH & 87.57\%& 5.7(M)& 23(h)\\
\hline
NASBOT& 87.70\%& 7.3(M)& 20(h)\\
\hline
Auto-Pytorch & 90.00\%& 8.2(M)& 10(h)\\
\hline
Auto-Keras & 88.56\% & 4.3(M)& 12(h)\\
\hline
 Ours (EfficientDL) & 91.31\%& 6.5(M)& 0.5(h)\\
 \hline
\end{tabular}
\end{table}

\subsection{Recommended key Components in Existing DL models }\label{sec:existing_DL}

We focus on evaluating the framework's ability to recommend key components of existing model models, as models proposed in various studies may not always reflect their optimal performance potential. 
% Next, we will provide a detailed introduction to the experiment and results.

\subsubsection{Baselines for existing DL architectures}

We evaluate our EfficientDL's effectiveness in recommending other critical parameters beyond the following model architecture:

\smallskip\noindent\textbf{ResNet50} ~\cite{He2016cvpr} is a type of Convolution Neural Network introducing the concept of residual learning, where the network learns residual mappings instead of learning unreferenced functions. Meanwhile, as a widely recognized backbone in image-related tasks, ResNet is selected as one of the models for our experiments.

\smallskip\noindent\textbf{MobilenetV3} ~\cite{Andrew2019iccv} is a CNN model designed for mobile and embedded vision applications, offering a balance between accuracy and efficiency.  Given its significant advancements in the development of lightweight neural networks, we select MobileNetV3 as one of the existing DL architectures for our study.

\smallskip\noindent\textbf{EfficientNet-B0} ~\cite{Tan2020icml} is highly effective for transfer learning, making it a popular choice for various vision tasks beyond image classification, such as object detection, segmentation, and more.

\smallskip\noindent\textbf{MaxViT-T}~\cite{tu2022maxvit} combines the strengths of CNNs and Vision Transformers (ViTs), which leverages convolutional layers to capture local patterns and attention mechanisms from transformers to model long-range dependencies. MaxViT-T is suitable for a wide range of computer vision tasks, especially in environments with limited computational resources.

\smallskip\noindent\textbf{Swin-B}~\cite{liu2021swin} is a powerful and versatile vision transformer that addresses the challenges of handling high-resolution images and achieving efficient computation. Its hierarchical structure, combined with the innovative shifted window attention mechanism, enables it to perform exceptionally well across a wide range of computer vision tasks, making it a valuable tool in both research and practical applications.

\smallskip\noindent\textbf{DaViT-T}~\cite{ding2022davit} is a significant advancement in vision transformer design, combining the strengths of dual attention mechanisms with a compact, efficient architecture.

\subsubsection{Implementation Details}

Following the existing DL architectures, we run our compared DL architectures on the ImageNet~\cite{Jia2009cvpr} dataset for image classification task. Specifically, we first use $\alpha\beta$-BO search method to recommend the components of epoch, batch size, and data augmentation for these models (i.e., ResNet50, MobilenetV3, EfficientNet-B0, MaxViT-T, Swin-B, and DaViT-T). Subsequently, each model architecture is trained with the recommended parameters on a single Nvidia GeForce RTX 3090 GPU. We search these components since these three components rank among the Top-10 key components aside from model architecture and data dimensions in Section~\ref{sec:component_recommendation}. We report the Top-1 accuracy on the test set for all experiments. 

\subsubsection{ Experimental Analysis.}

\smallskip\noindent\textbf{Our EfficientDL offers more efficient parameter recommendations for existing model architectures, enabling them to achieve superior performance, which demonstrates the potential to optimize even well-established models beyond their original capabilities.} Table~\ref{tab:existing_DL_result} shows the Top-1 accuracy of existing DL architectures with our recommended parameters on test set of ImageNet. Based on the observations from Table~\ref{tab:existing_DL_result}, the model architecture reported in the original paper does not necessarily yield the best performance. By fine-tuning parameters such as data augmentation, the number of iterations, and batch size, the performance can surpass that of alternative architectures by 0.12\% to 0.69\%. Notably, the MaxViT-T model shows significant improvement after just 290 iterations, while the other five comparison models require longer training times to achieve their optimal performance.

More importantly, the static performance prediction model in our EfficientDL could almost precisely predict the performance of a DL system under given parameters. From Table~\ref{tab:existing_DL_result}, we can observe that the error range of Top-1 accuracy between prediction and running is merely from 0.01\% to 0.09\%, which highlights that the static performance prediction model, built on our multi-dimensional system component dataset, endows EfficientDL with the accuracy ability to foresee outcomes.

\begin{table}
    \centering
    \caption{Results of existing DL models with our recommended parameters on ImageNet. Noted that in the 'X/Y' values of the 'Our' row in the 'Top-1 accuracy' column, 'X' represents the performance achieved using the recommended parameters through actual runs, while 'Y' represents the result predicted by our static performance prediction model.}
    \label{tab:existing_DL_result}
    \small
    \scalebox{0.85}{
    \begin{tabular}{p{0.12\textwidth}p{0.042\textwidth}p{0.026\textwidth}p{0.15\textwidth}p{0.03\textwidth}p{0.065\textwidth}}
    \toprule
    \textbf{Model}  & \textbf{Type} & \textbf{Epoch} & \textbf{Data Augmentation}    & \textbf{Batch size}  & \textbf{Top-1 accuracy (\%)}   \\ 
    \midrule
    \multirow{2}{*}{\textbf{ResNet50}} &Original& 300   & CutMix, ColorJitter, Random resize scale, Lighting, horizontal flip  & 256                         & 78.32  \\
    \cline{2-6}
    &  Our   & 390   & horizontal flip, random resized crop, ColorJitter, mixup, cutmix  & 192 & \textbf{78.57}/78.61\\ 
    \midrule
    \multirow{2}{*}{\textbf{MobilenetV3}} &Original & 300   & AutoAugment   & 4096     & 75.20\\
    \cline{2-6}
    & Our & 450   & random resized crop, Random erase, mixup & 256 & \textbf{75.43}/75.48  \\
    \midrule
    \multirow{2}{*}{\textbf{EfficientNet-B0}}  &Original  & 350   & AutoAugment   & 2048                         & 77.11\\
    \cline{2-6}
    &Our  & 450   & AutoAugment, Random resize scale, Random resize aspect ratio, horizontal flip, Color jitter, Random erase, mixup & 512 & \textbf{77.80}/77.82\\
    \midrule
    \multirow{2}{*}{\textbf{MaxViT-T}}& Original & 300   & Center crop, RandAugment, Mixup & 4096             & 83.62\\
    \cline{2-6}
    &Our  & 290   & Random resize scale, Random resize aspect ratio, horizontal flip, color jitter, mixup, cutmix, Random erase & 128  & \textbf{84.01}/83.92\\
    \midrule
    \multirow{2}{*}{\textbf{Swin-B}} & Original  & 300   &AutoAugment,color~jitter,cutmix,mixup,Random erase  & 1024  & 83.11\\
    \cline{2-6}
    &Our            & 400   & AutoAugment, horizontal flip, mixup, color~jitter, Random erase    & 256  & \textbf{83.24}/83.23\\
    \midrule
    \multirow{2}{*}{\textbf{DaViT-T}} & Original   & 300   & Random resize scale, horizontal flip, Color jitter, AutoAugment, Random erase, mixup, cutmix  & 2048   & 82.12\\
    \cline{2-6}
    &Our   & 340   & horizontal flip, vertical flip, Random resize aspect ratio, Random resize scale, Color jitter, AutoAugment, Random erase, mixup, cutmix & 128 & \textbf{82.34}/82.27     \\
    \bottomrule
    \end{tabular}}
\end{table}

\subsection{Ablation study}

Our EfficientDL is composed of multiple elements, including specially designed system components, importance-based component recommendations, and our proposed enhanced $\alpha\beta$-BO search method. In this section, we evaluate the contributions of each element to the overall effectiveness of the framework, demonstrating their collective impact on performance optimization.

\subsubsection{Baselines for ablation study}

We conduct ablation experiments to comprehensively analyze the impact of various elements within our EfficientDL, with a particular focus on the following key aspects:

\smallskip\noindent\textbf{Ablation experiments focused on our comprehensive fine-grained components} are designed to evaluate whether each component is essential and fully operational within our EfficientDL. Specifically, we randomly exclude three, six, and nine components, respectively, to create different versions of our framework. We then generate model parameter recommendations based on these modified frameworks and run the models using the recommended parameters to compare the performance of DL models.

\smallskip\noindent\textbf{Ablation experiments focused on key component recommendations} are primarily aimed at evaluating whether these recommended components have a decisive impact on the performance of DL models. Specifically, we randomly select other components except for the key components as the basis for recent recommendations, and then run the models using the recommended parameters to compare the performance of DL models.

\smallskip\noindent\textbf{Ablation experiments focused on our proposed $\alpha\beta$-BO search method} are designed to evaluate the effectiveness of our strategy at its optimal parameters. Specifically, we compare our $\alpha\beta$-BO search method with Bayesian optimization methods based on Expected Improvement (EI)~\cite{Helen2020ES}, Probability of Improvement (PI)~\cite{Carolin2022NeurIPS}, and Upper Confidence Bound (UCB)~\cite{Maryam2020MLKDD}, as well as $\alpha\beta$-BO search method without probability parameter. We use these methods to recommend parameters for the model and then run the model with the recommended parameters to compare the performance of DL models.

\subsubsection{Implementation Details}

We focus the ablation experiments on existing model architectures, including ResNet50, MobileNetV3-Large, and Swin-B on the ImageNet dataset for image classification task. We apply different versions of our framework to recommend the parameters of the existing model architectures. We then train each compared DL architecture with the recommend parameters on a single GPU (Nvidia GeForce RTX 3090). For ablation experiments that focused on
comprehensive fine-grained components and $\alpha\beta$-BO search method, we apply these different versions of our framework to recommend the components of epoch, batch size, and data augmentation, similar to Section~\ref{sec:existing_DL}. For the ablation experiments focused on key component recommendations, we apply the different versions of our framework to recommend the components of Learning Rate Schedules, learning rate and Optimization algorithm. We report the Top-1 accuracy on the test set for all experiments.

\subsubsection{Experimental Analysis}

\smallskip\noindent\textbf{Elements including specially designed system components, importance-based component recommendations, and our proposed enhanced $\alpha\beta$-BO search method within our EfficientDL synergistically contribute to the optimal recommendation of deep learning models, highlighting their integral role in enhancing model performance.} Table~\ref{table:ablation} shows the results of ablation study on different elements included in our EfficientDL. From Table~\ref{table:ablation}, we observe the following: 

\begin{itemize}[wide = 0pt, itemsep = 3pt]

\item When randomly removing 3, 6, or 9 system components, the performance of the DL model trained with parameters recommended by our framework declines from 0.07\% to 2.02\%.  Meanwhile, the predictive performance of the random forest regression model also decreases from 0.00118 to 0.00054 after random component removal, highlighting the comprehensiveness of the system components we designed for our EfficientDL.

\item When our model randomly recommends parameters for non-essential components, the DL model fails to achieve the performance of models with key components, underscoring the importance of focusing on key components during parameter recommendation.

\item DL model trained with parameters identified by our proposed enhanced $\alpha\beta$-BO search achieves the best performance, demonstrating the robustness of our approach in avoiding local optima and ensuring optimal outcomes.

\end{itemize}

\begin{table*}[tb]
\caption{Results of ablation study on different elements included in EfficientDL}\label{table:ablation}
\begin{tabular}{p{0.10\textwidth}p{0.14\textwidth}p{0.06\textwidth}p{0.36\textwidth}p{0.1\textwidth}p{0.13\textwidth}} 
\hline
\textbf{Model} & \textbf{Method} &\textbf{Epoch} &\textbf{Data Augmentation}&\textbf{Batch size} &  \textbf{Top-1 accuracy (\%)}\\
\hline
\multirow{8}{*}{\textbf{ResNet50}} &EfficientDL (Remove three components) &375&horizontal flip, RandomRotation, mixup, cutmix & 192 & 77.03\\
\cline{2-6}
& EfficientDL (Remove six components) &470& horizontal flip, RandomRotation, GaussianBlur, mixup & 96 & 76.95\\
\cline{2-6}
& EfficientDL (Remove nine components) & 405& vertical flip, horizontal flip, GaussianBlur, mixup & 50& 76.85\\
\cline{2-6}
& EfficientDL\_EI &335&horizontal flip, RandomRotation, GaussianBlur, mixup & 192&77.18\\
\cline{2-6}
& EfficientDL\_PI &390& horizontal flip, Vertical flip, mixup & 400 &77.01\\
\cline{2-6}
& EfficientDL\_UCB &440&horizontal flip, Random erase, cutmix&256&77.19\\
\cline{2-6}
& EfficientDL\_noprobability &380&horizontal flip, Random resize scale, Random resize scale, mixup & 224& 76.94\\
\cline{2-6}
& EfficientDL & 390   & horizontal flip, random resized crop, ColorJitter, mixup, cutmix  & 192 & \textbf{78.57}\\
\hline
\multirow{8}{*}{\textbf{MobilenetV3}} &EfficientDL (Remove three components) &400&horizontal flip, AutoAugment, mixup&440&75.33\\
\cline{2-6}
& EfficientDL (Remove six components) & 360& RandomRotation, Random erase&512&75.30\\
\cline{2-6}
& EfficientDL (Remove nine components) & 380& horizontal flip, Vertical flip, Random erase, mixup &400&74.70\\
\cline{2-6}
& EfficientDL\_EI &385&horizontal flip, AutoAugment&256&75.22 \\
\cline{2-6}
& EfficientDL\_PI & 390& random resized crop, AutoAugment&192&74.78\\
\cline{2-6}
& EfficientDL\_UCB &395&random resized crop, AutoAugment&384&75.18\\
\cline{2-6}
& EfficientDL\_noprobability & 385&Random erase, mixup&512&75.36\\
\cline{2-6}
& EfficientDL& 450   & random resized crop, Random erase, mixup & 256 & \textbf{75.43}\\
\hline
\multirow{8}{*}{\textbf{Swin-B}} &EfficientDL (Remove three components) &480&horizontal flip, Random erase, AutoAugment, Random resize aspect ratio &400&81.22\\
\cline{2-6}
& EfficientDL (Remove six components) &475&Random resize scale, horizontal flip, Random erase, mixup&512&83.01\\
\cline{2-6}
& EfficientDL (Remove nine components) &345&horizontal flip, Vertical flip, color jitter&512&82.23\\
\cline{2-6}
& EfficientDL\_EI &425&horizontal flip, Random erase, AutoAugment, mixup&192&82.56
\\
\cline{2-6}
& EfficientDL\_PI &390&horizontal flip, Vertical flip, AutoAugment, mixup&50&81.57\\
\cline{2-6}
& EfficientDL\_UCB &390&horizontal flip, GaussianBlur, AutoAugment, mixup&90&82.20\\
\cline{2-6}
& EfficientDL\_noprobability &395&Vertical flip, color jitter, Random erase, AutoAugment &90&82.86\\
\cline{2-6}
& EfficientDL & 400   & AutoAugment, horizontal flip, mixup, color~jitter, Random erase    & 256  & \textbf{83.24}\\
\hline
%\textbf{Model} & \textbf{Method} &\textbf{learning rate} &\textbf{Learning Rate Schedules}&\textbf{Optimization algorithm} &  \textbf{Top-1 accuracy (\%)}\\
%\hline
\textbf{ResNet50}&EfficientDL &0.05&Step Decay&SGD&78.47\\
\cline{1-1}\cline{3-6}
\textbf{MobilenetV3}&(Without key  &0.005&Cosine&Adam&73.75\\
\cline{1-1}\cline{3-6}
\textbf{Swin-B}&components)&0.25&Step Decay&SGD&75.98\\
\hline
\end{tabular}
\end{table*}

\subsection{Downstream Object Detection}

The utilization of models (e.g., Swin-B or MobileNetV3) pre-trained on image classification tasks as backbones for downstream tasks (e.g., object detection) is a prevalent practice.  Therefore, we evaluate whether the DL models trained with our recommended parameters can effectively serve as backbones for downstream tasks, thereby further enhancing their performance.

\subsubsection{Baselines for Downstream tasks}

We conduct experiments to analyze the performance of the DL models trained with the recommended parameters as backbones in object detection tasks, focusing primarily on the following two methods:

\smallskip\noindent\textbf{SSDLite} ~\cite{Andrew2019iccv} apply MobileNetV3 as a drop-in replacement for the backbone feature extractor to perform detection on the COCO dataset. Following the method by ~\cite{Andrew2019iccv}, the first layer of SSDLite is attached to the last feature extractor layer with an output stride of 16, and the second layer is attached to the last feature extractor layer with an output stride of 32.

\smallskip\noindent\textbf{Cascade Mask R-CNN} ~\cite{Cai2018CVP} is a multi-stage object detection architecture where deeper stages in the cascade become progressively more selective against close false positives. Following the method by~\cite{liu2021swin}, Swin-B is integrated into the Cascade Mask R-CNN framework to perform detection on the COCO dataset. 

\subsubsection{Implementation Details}

Following~\cite{liu2021swin, Andrew2019iccv}, we apply MobileNetV3 and Swin-B trained with our recommended parameters in Section~\ref{sec:existing_DL} as the backbones of SSDLite and Cascade Mask R-CNN. We train SSDLite and Cascade Mask R-CNN on a single GPU (Nvidia GeForce RTX 3090) with the COCO dataset. We report the mAP on the test set for all experiments.

\subsubsection{Experimental Analysis}

\smallskip\noindent\textbf{Models trained with the parameters recommended by our EfficientDL framework can effectively serve as the backbone for downstream tasks, significantly boosting their performance and demonstrating the robustness of our approach.} 
From Table~\ref{tab:object_detection}, we observe that SSDLite combined with our recommended pre-trained MobileNetV3 improves the performance of object detection from 0.22 to 0.222, while Cascade Mask R-CNN with our recommended pre-trained swin-B improves the performance of object detection from 0.449 to 0.459.

\begin{table}
    \centering
    \caption{Results of compared object detection model with our trained models as the backbone  on COCO dataset}
    \label{tab:object_detection}
    \begin{tabular}{p{0.15\textwidth}p{0.15\textwidth}p{0.1\textwidth}}
    \toprule
    \textbf{Method}  & \textbf{Backbone} & \textbf{mAP}\\ 
    \midrule
    \multirow{2}{*}{\textbf{SSDlite}} &MobileNetV3\_Original&0.22  \\
    \cline{2-3}
    & MobileNetV3\_Our    &\textbf{0.222}\\   
    \hline
    \multirow{2}{*}{\textbf{Cascade Mask R-CNN}}  &swin-B\_Original  &0.449\\
    \cline{2-3}
    &swin-B\_Our  &\textbf{0.459}\\
    \hline
    \end{tabular}
\end{table}

The promising results achieved with EfficientDL underscore its significant potential to streamline and optimize the development of deep learning systems across various domains. By enabling rapid and reliable system configuration, our framework allows practitioners to concentrate on high-level design considerations and innovative applications rather than on tedious manual tuning. 

%% file: 07-Conclusion.tex
\section{Conclusion}\label{sec:conclusion}

%In this work, we present EfficientDL, an innovative deep-learning board designed for automatic performance prediction and component recommendation. Unlike existing AutoDL frameworks, which rely on training feedback from actual runs, EfficientDL  can quickly and precisely recommend twenty-seven system components and predict the performance of DL models benefiting from our proposed comprehensive, multi-dimensional, fine-grained system component dataset. Such dataset ensable us to develop a static performance prediction model and comprehensive optimized component recommendation algorithm (i.e., $\alpha\beta$-BO search) in a data-driven manner, removing the dependency of EfficientDL on actually running parameterized models during the traditional optimization search process.

%without requiring any
%training feedback.

%provides a comprehensive solution by incorporating 27 system components—ranging from pooling operations and skip connections to data augmentation techniques and hardware configurations like GPU devices. This broad coverage allows our framework to deliver accurate performance predictions swiftly, without the need for iterative training feedback, significantly reducing the time and resources typically required for manual system design.

In this study, we propose EfficientDL, an innovative deep learning board designed for automatic performance prediction and component recommendation. Unlike existing AutoDL frameworks that rely on training feedback from actual runs, EfficientDL can swiftly and accurately recommend 27 system components and predict DL model performance, enabled by our comprehensive, multi-dimensional, fine-grained system component dataset. This dataset enables the development of a static performance prediction model and a highly optimized component recommendation algorithm (i.e.,$\alpha\beta$-BO search) in a data-driven manner, eliminating the need for EfficientDL to run parameterized models during the traditional optimization process.

Our experimental evaluations on image classification tasks, particularly with the CIFAR-10 dataset, demonstrate that EfficientDL surpasses existing AutoML frameworks in both accuracy and efficiency. Meanwhile, the simplicity and power of EfficientDL stem from its compatibility with most DL models. For example, EfficientDL delivers competitive performance improvements across six widely-used models, including ResNet50, MobileNetV3, EfficientNet-B0, MaxViT-T, Swin-B, and DaViT-T. Furthermore, when employing the pre-trained models generated by our recommended components as backbones for object detection tasks, EfficientDL improves the mean Average Precision (mAP) by more than 0.49 for Swin-B and 0.22 for MobileNetV3.